\documentclass[journal]{IEEEtran}

\usepackage{amsmath,amsfonts}
\usepackage{array}
\usepackage[caption=false,font=normalsize,labelfont=sf,textfont=sf]{subfig}
\usepackage{textcomp}
\usepackage{stfloats}
\usepackage{url}
\usepackage{verbatim}
\usepackage{graphicx}
\def\BibTeX{{\rm B\kern-.05em{\sc i\kern-.025em b}\kern-.08em
    T\kern-.1667em\lower.7ex\hbox{E}\kern-.125emX}}
\usepackage{balance}

\usepackage{cite}

\usepackage{algorithm}
\usepackage{algpseudocode}

\usepackage{color, xcolor}
\definecolor{revised}{RGB}{0,0,0}

\usepackage{hyperref}
\usepackage[capitalise]{cleveref}
\usepackage{siunitx}
\usepackage{booktabs}
\usepackage{multirow}

\newcommand{\ieeefellow}{{\IEEEmembership{Fellow, IEEE}}}
\newcommand{\ieeemember}{{\IEEEmembership{Member, IEEE}}}

\newcommand{\ieeegraduatestudentmember}{{\IEEEmembership{Graduate Student Member, IEEE}}}

\newcommand{\ie}{\textit{i.e.}}

\begin{document}

\title{
{\color{revised}Egocentric} Station Holding of Robotic Fish in Unknown Turbulent Background Flow
}

\author{Xiaozhu Lin, \ieeegraduatestudentmember, Xu Huang, Hongru Dai, Xiaopei Liu, \\ Junzhi Yu, \ieeefellow, and Yang Wang, \ieeemember%
\vspace{-0.5cm}
\thanks{This work was supported by the National Natural Science Foundation of China under Grant 62503329. \textit{(Corresponding author: Yang Wang.)}}
\thanks{Xiaozhu Lin, Xu Huang, Hongru Dai, Xiaopei Liu and Yang Wang are with the School of Information Science and Technology, ShanghaiTech University, Shanghai, China {\tt\small \{linxzh, huangxu2024, daihr2024, liuxp, wangyang4\}@shanghaitech.edu.cn}}
\thanks{Junzhi Yu is with the State Key Laboratory for Turbulence and Complex Systems, Department of Advanced Manufacturing and Robotics, College of Engineering, Peking University, Beijing 100871, China, and also with the Laboratory of Cognitive and Decision Intelligence for Complex System, Institute of Automation, Chinese Academy of Sciences, Beijing 100190, China {\tt\small junzhi.yu@ia.ac.cn}}
}

\markboth{Journal of \LaTeX\ Class Files,~Vol.~18, No.~9, September~2020}{\MakeLowercase{\textit{(et al.)}}: How to Use the IEEEtran \LaTeX \ Templates}

\maketitle


\begin{abstract}
Approaching a target position and holding station in flowing water is a fundamental and critical capability for robotic fish operating in natural aquatic environments. 
\textcolor{revised}{Despite decades of advances in enhancing swimming efficiency and maneuverability, this capability remains underdeveloped—largely owing to the insufficiently characterized, highly nonlinear fluid–structure interactions inherent to freely swimming robotic fish in flows.}
\textcolor{revised}{
To bridge this gap, we propose the SWiFT framework, a Swimming With Flow Toolbox that enables the efficient exploration of an egocentric  station-holding policy for a body and/or caudal fin (BCF) robotic fish in unknown and turbulent background flows via reinforcement learning (RL).}
Our SWiFT integrates a free-swimming flow-tank experimental setup with a highly efficient, physically consistent Computational Fluid Dynamics (CFD)-based simulator and a systematic sim-to-real transfer pipeline. The resulting policy achieves substantial improvements over state-of-the-art methods across all metrics, most notably root‑mean‑square error (RMSE) of distance.
Furthermore, we validated that \textcolor{revised}{egocentric} feedback alone, without any explicit flow sensing, enables station-holding in unknown turbulent flows, closely mirroring the biological phenomenon of rheotaxis. 
\textcolor{revised}{Accordingly, the success of this egocentric station-holding policy not only advances robotic fish control toward real-world deployment, but also highlights SWiFT’s promise as a foundation for tackling complex swimming tasks for underwater robots.}

\end{abstract}


\section{Introduction}\label{sec:intro}

\begin{figure}[tb]
    \centering
    \includegraphics[width=0.99\linewidth]{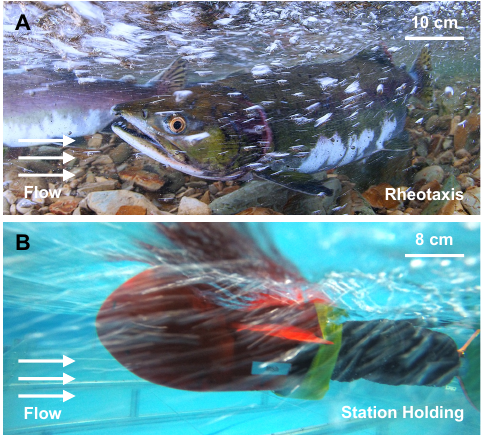}
    \caption{\textbf{Comparison of biological fish \emph{rheotaxis} and robotic fish station holding.} (\textbf{A}) A biological fish performing \emph{rheotaxis} behavior in a turbulent stream. (\textbf{B}) Our bio-mimetic robotic fish performing station-holding capability in turbulent background flow, successfully replicating a fundamental biological behavior to resist currents with lifelike agility and stability. (\textbf{A}) is adopted from https://www.flickr.com/photos/torchuck/15058915295/.}
    \label{fig:hook}
\end{figure}

\IEEEPARstart{E}{nvironmentally} adaptable operation in natural aquatic environments, characterized by high energy efficiency \cite{yan2020efficient,li2024towards} and agile maneuverability \cite{lin2025learning,yu2025three}, remains a key motivation for developing bio-inspired underwater robots \cite{yan2024recent}.
To this end, robotic fish propelled by BCF have delivered notable progress over the past decade, with significant advancements in electromechanical design \cite{zhong2017novel,wang2018three,yan2023towards}, underwater perception \cite{zheng2020online,zhang2023real}, and locomotion control \cite{yan2020efficient,zhang2022simulation,zhong2023general}, etc.
However, these progress have not readily translated to natural rivers and oceans. A primary obstacle is that existing research focuses predominantly on quiescent water \cite{yan2020efficient,zhang2021decentralized,zhang2022simulation}, largely neglecting the unknown, commonly turbulent background flows prevalent in real-world environments. These flows interact with the robot, creating a complex dynamic field that exerts nonlinear and time-varying forces and moments. Consequently, most control paradigms—centered on attitude, set-point, or trajectory tracking \cite{zhang2020path, zhang2022leveraging, lin2025learning}—fail to account for the critical influence of background flow.
\textcolor{revised}{More crucially, station-holding, defined as the capability to freely swim to a target position and subsequently maintain it against unknown background flow disturbances (see \cref{fig:hook}), is an essential requirement for executing underwater tasks \cite{whitcomb2000underwater}. However, such a capability of a BCF robotic-fish has received insufficient systematic investigation, despite being both a pivotal research frontier and a fundamental prerequisite for reliable autonomy in natural aquatic settings \cite{jung2013flow,wang2024investigation,liu2025artificial}.}

For mainstream multi-joint BCF-propelled robotic fish \cite{li2019bottom,yan2020efficient,zhang2022simulation}, the difficulty of achieving stable station-holding in unknown, turbulent flows can be attributed to three main factors: 
First, fluid-structure interactions (FSIs) between the oscillatory body of the robotic fish and the surrounding flows lead to nonlinear and time-varying hydrodynamic surface forces \cite{borazjani2008numerical}. 
These forces are tightly coupled with the posture of the robotic fish and the inflow conditions.
Second, from a control perspective, robotic fish are inherently underactuated systems \cite{yu2016data,li2019bottom,zhang2022simulation}. 
Station-holding imposes particularly stringent requirements on such systems, demanding not only precise positioning but also sustained rejection of flow-induced disturbances.  
Third, the lack of prior flow information and specialized flow-sensing capabilities means the robot must rely on real-time \textcolor{revised}{egocentric} measurements to infer and counteract unknown flow disturbances.  
Collectively, these difficulties pose unique challenges for conventional methods\cite{wang2015averaging,yu2016data,li2019bottom}, which rely on simple dynamic models, extensive parameter identification, and specialized tuning. As such, these established approaches prove difficult to apply directly to the problem with unknown turbulent background flow.

\textcolor{revised}{Despite the considerable challenges for robotic fish swimming in background flows, real fish's station-holding is a typical widespread biological  \emph{rheotaxis} behavior \cite{coombs2020rheotaxis}\cite{liao2006role}. }
This phenomenon has inspired some pioneering robotic-fish studies in flowing water over the past decade; however, these studies have primarily been conducted in confined flow tunnels where the robot is usually partially restrained \cite{liu2025artificial,zhang2015distributed,jung2013flow,salumae2013flow}.  
While these flow tunnels provide advantages including precise flow control, repeatable hydrodynamic conditions, and real-time force/torque measurement capabilities, their confined space and simplified flow environments inherently limit the scope of investigation.
These limitations result in an oversimplification of FSI and prevent the robot from exploiting its full motion potential for flow counteraction, such as the tail-to-flow maneuvering \cite{zhang2015distributed,jung2013flow,salumae2013flow}.  
Alternatively, inspired by the lateral line system \cite{zhai2021fish}, some studies have outfitted robotic fish with the artificial lateral line system (ALLS) to enhance flow perception and control performance \cite{jung2013flow,salumae2013flow,salumae2012against}. 
However, ALLS can be easily corrupted by self-induced turbulence and sensor noise, resulting additional calibration and modeling efforts \cite{jung2013flow}.
In summary, control strategies developed for confined, simplified flows or perception-reliant scenarios struggle to generalize to unknown, turbulent environments \cite{wang2024investigation}. 
Compellingly, biological studies have demonstrated that even fish with impaired lateral line systems retain flow-sensing and orientation capabilities \cite{liao2006role}. \textcolor{revised}{
This finding prompts a fundamental question: \emph{Can robotic fish achieve stable station holding in unknown flow fields using egocentric information, without explicit flow-related measurements?}
Yet, despite this biological precedent, replicating such \textcolor{revised}{egocentric} \emph{rheotaxis} behavior in robotic fish is rather challenging \cite{liu2025artificial}. }

\textcolor{revised}{
To address this challenging, we propose a novel framework for studying the station holding of robotic fish swimming freely in flowing water, which we term SWiFT (Swimming With Flow Toolbox). SWiFT couples a free-swimming flow tank with an efficient LBM simulator (GPU-accelerated, physics-based, low numerical dissipation \cite{li2020fast,lee2018dart,woodward1988skinning}) via a systematic sim-to-real transfer framework.} 
\textcolor{revised}{Furthermore, SWiFT implements a soft actor–critic (SAC) algorithm \cite{haarnoja2018soft}, incorporating a carefully engineered reward function and comprehensive domain randomization to enhance policy robustness. Departing from conventional pipelines, this end-to-end RL approach eliminates the need for low-level periodic controllers (e.g., CPG) and instead directly maps sensory inputs to joint-angle commands, thereby maximizing agility. This formulation reveals the critical role of non-periodic body kinematics in achieving robust station holding. Crucially, SWiFT is not merely an assembly of modules but an integrated framework forged through a coherent chain of design innovations. The inherent complexity of station holding motivates a model-free paradigm, guiding the adoption of RL \cite{song2023reaching}. In turn, the data-intensive demands of RL necessitate a GPU-accelerated low numerical dissipation LBM solver \cite{lee2018dart, mittal2025isaac} and deliberate efficiency trade-offs realized through a suite of sim-to-real techniques \cite{peng2018sim, tan2018sim}. Central to this integration, the sim-to-real transfer pipeline calibrates the CFD simulator using only a limited amount of experimental data from the flow tank. The resulting calibrated environment strikes an explicit trade-off between hydrodynamic fidelity and computational efficiency, preserving essential fluid–structure interaction (FSI) nonlinearities while remaining tractable for data-intensive RL workflows. }

\textcolor{revised}{Extensive real-world comparison experiments demonstrate that the proposed SWiFT framework enables, for the first time, robust egocentric station holding of a robotic fish in unknown, turbulent background flows. Across all key performance metrics, including success rate, root‑mean‑square error (RMSE), and approaching time, our method consistently outperforms state-of-the-art baselines\cite{guelman2007qualitative,li2019bottom,wang2025identification,zhang2022simulation,mamakoukas2021derivative}, ranging from model-based model predictive control (MPC) to data-driven Koopman operator approaches. Beyond advancing robotic fish toward practical outdoor deployment, these results substantiate an important observations in real fish: rheotaxis can be achieved solely through egocentric feedback, without requiring explicit flow sensing. Moreover, the success of SWiFT underscores the considerable potential of coupling reinforcement learning (RL) with CFD for complex underwater locomotion control, which serves as one of our core methodological contributions. While the data-intensive nature of RL and the prohibitive cost of CFD solvers have historically discouraged such endeavors, our work reveals that this barrier can be overcome by exploiting the intrinsic parallelism and low numerical dissipation of LBM \cite{song2025creating,liu2025hybrid}, the inherent robustness of RL policies, and the favorable accuracy–efficiency trade-off unlocked via minimal real-to-sim calibration. By tightly integrating these elements, we successfully derive an end-to-end, egocentric swimming policy for robotic fish that capable of robust station holding in unknown and turbulent background flow.}

Our main contributions are summarized as follows:

\begin{enumerate}
    \item \textcolor{revised}{\textit{First Formal Problem Formulation and Solution.} To the best of our knowledge, this work provides the first systematic formulation, analysis, and experimental validation of egocentric station holding for a freely swimming robotic fish in unknown, turbulent background flows—a capability that is both biologically ubiquitous and critically lacking in existing robotic fish study.}
    \item \textcolor{revised}{\textit{High-Performance, Robust Control Policy.} We derive an end-to-end, egocentric swimming policy that consistently outperforms SOTA methods, including model-based MPC and data-driven Koopman approaches, across all key metrics (success rate, RMSE, approaching time). The policy exhibits strong robustness to flow perturbations while maintaining a simple and efficient architecture.}
    \item \textcolor{revised}{\textit{An Efficient Sim-to-Real Framework Enabling CFD–RL Integration.} SWiFT identifies a distinct pathway for efficient control policy generation by successfully bridging LBM and RL. By exploiting the intrinsic parallelism and low numerical dissipation of LBM alongside a data-efficient real-to-sim calibration strategy, SWiFT overcomes the historical barrier between CFD and data-intensive learning. This integration ultimately enables the zero-shot transfer of the learned policy to a robotic fish in a free-swimming flow tank.}
    \item \textcolor{revised}{\textit{Fundamental Insight and Practical Pathway Toward Natural Waterways.} This study establishes a key ethological and engineering insight: stable station holding can be achieved solely through egocentric feedback, without any explicit flow-field perception—a finding that aligns closely with biological evidence on rheotaxis in fish. Collectively, these advances lay a critical foundation for deploying robotic fish in unstructured, real-world aquatic environments.}
\end{enumerate}


\begin{figure*}[tb]
    \centering
    \includegraphics[width=0.99\linewidth]{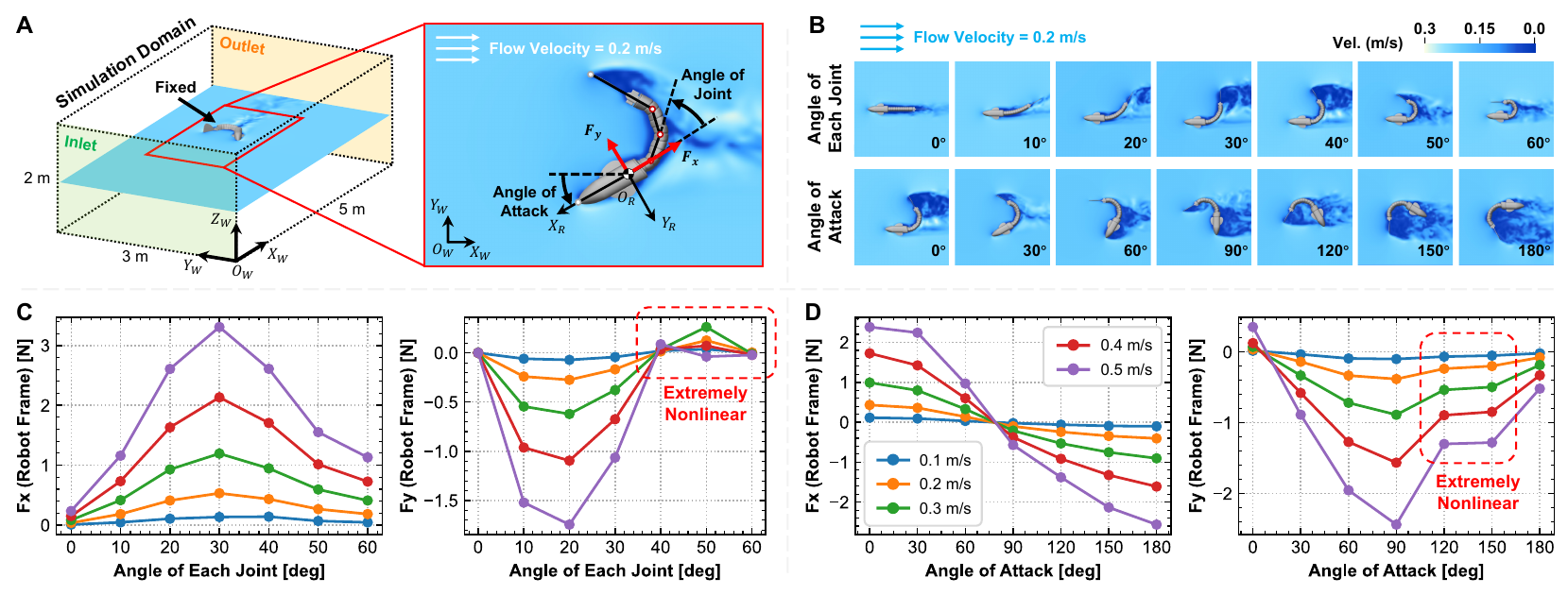}
    \vspace{-10pt}
    \caption{\textbf{Characterization of hydrodynamic forces on a robotic fish in background flow using the CFD-based simulation.} (\textbf{A}) Simulation setup (5~m $\times$ 3~m $\times$ 2~m) featuring inlet-outlet boundary conditions and coordinate frames (world and robot). The robot’s posture is defined by two key parameters: the AoA (between background flow and the head segment) and the Angle of Joint. The latter—set uniformly across all joints—defines the robot’s overall curved body morphology. (\textbf{B}) Flow velocity field visualization at 0.2~m/s background flow: the first row shows fields with increasing joint angles (0°--~60°) at a fixed AoA (0°), and the second row shows fields with increasing angles of attack (0°--~180°) at a fixed joint angle (40°). Background flow direction, velocity magnitude, and colormap are indicated. (\textbf{C}) Hydrodynamic force components (drag: robot frame’s negative X direction; lateral force: robot frame’s negative Y direction) as functions of joint angle across five background flow velocities (0.1--0.5~m/s). (\textbf{D}) Hydrodynamic force components as functions of AoA across the same background flow velocity range.}
    \label{fig:flow_impact}
\end{figure*}

\begin{table*}[t]
	\centering
	\caption{Review of Representative Works on Robotic Fish in Background Flow Field}
	\includegraphics[width=0.99\linewidth]{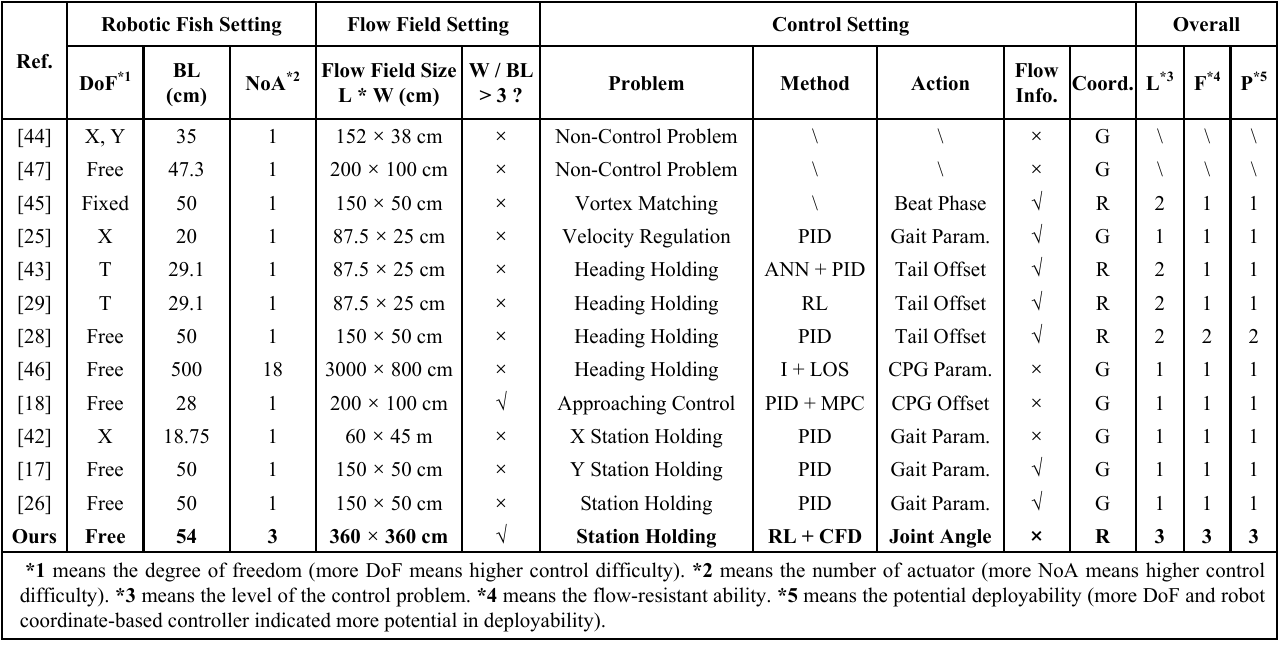}
	\label{tab:related_work}
\end{table*}

\section{Related Works}\label{sec:related}

\subsection{The Nonlinearities of Fluid-Structure Interaction}
Unlike quiescent water conditions, where hydrodynamic forces can be partially approximated using simplified models, turbulent background flows introduce spatiotemporally varying perturbations that significantly amplify the complexity of FSIs \cite{jung2013flow}. Specifically, the oscillatory body motions of a BCF robotic fish interact dynamically with incoming currents, leading to time-varying hydrodynamic surface forces that are highly sensitive to the robot's posture and inflow characteristics. 
To preliminarily glimpse into the inherent complexity FSIs under turbulent flows, we employ a high-fidelity CFD simulator to study the relationship between hydrodynamic forces (drag and lateral components) and key parameters such as joint angle (representing body flexion) and AoA (representing orientation relative to flow) under different background flow velocities. As illustrated in \cref{fig:flow_impact}, the results demonstrate that even under steady-state conditions with a fixed posture, hydrodynamic forces exhibit strong nonlinear dependencies on both parameters. 
This nonlinear relationship becomes even more complex when the fish starts swimming, as unsteady motions introduce additional temporal dynamics. This complexity highlights the necessity of high-fidelity CFD simulator and ultimately motivates the use of model-free RL in this work.
\subsection{Robotic Fish Control in Background Flow}

As shown in \cref{tab:related_work}, existing works on robotic fish control under background flows exhibit notable limitations in both the scope of addressed problems and achieved performance.  
On the one hand, most existing works focus on highly constrained scenarios \cite{zhou2010study,zhang2015distributed,zheng2020dual,zhong2021tunable}.  
Specifically, the authors of \cite{zheng2020dual,zheng2021learning} achieved AoA maintenance (tilt angle between the robotic fish and background flow) with only one rotational degree of freedom.  
In \cite{jung2013flow}, station holding was realized for a robotic fish limited to forward-backward motion.  
In \cite{jevzov2012sensing}, energy-efficient control was implemented for a fully fixed robotic fish in the wake of Karman vortex streets.  
Zhang et al. employed a narrow flow tunnel as a treadmill to investigate closed-loop velocity control of robotic fish in quasi-steady flow fields \cite{zhang2015distributed}.
The robots therein are typically restricted to very few degrees of freedom (DoF) or fully fixed, with control often limited to a single dimension.  
These simplifications, while rendering the control problem more tractable, ultimately limit the capability of the robot to swim against currents with full DoF.

\textcolor{revised}{
On the other hand, while a few studies addressed freely swimming robotic fish in flow fields \cite{wang2024investigation,jung2013flow,salumae2013flow,kelasidi2017integral,chen2021investigation}, they did not directly tackle the station-holding challenge in this work.} 
For instance, the authors of \cite{wang2024investigation} used a hybrid proportion-integration-differentiation (PID) with model predictive control (MPC) control approach to achieve robotic fish target approaching but did not realize station holding.  
In \cite{salumae2013flow}, the focus was on stabilizing robotic fish control solely along the y-axis.  
Furthermore, controllers designed by aforementioned methods \cite{wang2024investigation,jung2013flow,salumae2013flow} generally work in the world coordinate system, \textcolor{revised}{and the swimming strategies are constrained to a single flow speed.}
More importantly, the aforementioned studies assume the robotic fish always faces the incoming flow, without accounting for flow-facing-away scenarios. 
\textcolor{revised}{Our experiments (see \cref{fig:comparison_sota}, \cref{fig:comparison_qualitative} and \cref{fig:comparison_quantitative}) instead demonstrate that even with parameter tuning, the SOTA controllers lack robustness:} their performance degrades with flow velocity variations, and inconsistencies emerge across different target points even at the same velocity.

In summary, prior work primarily focuses on highly constrained settings, such as fixed orientations, reduced degrees of freedom, or unidirectional interactions with background flow, which significantly simplify the underlying dynamics and fail to capture the challenges of freely swimming robotic fish (see \cref{tab:related_work}).  
This necessitates the development of an experimental platform featuring a sufficiently large flow tank that avoids external fixtures or artificial constraints on degrees of freedom. Further, equipped with a programmable flow generator, such a platform finally enables the investigation of the practically relevant problem of station holding for a freely swimming robotic fish in background flow.

\begin{figure*}[tb]
    \centering
    \includegraphics[width=0.9\linewidth]{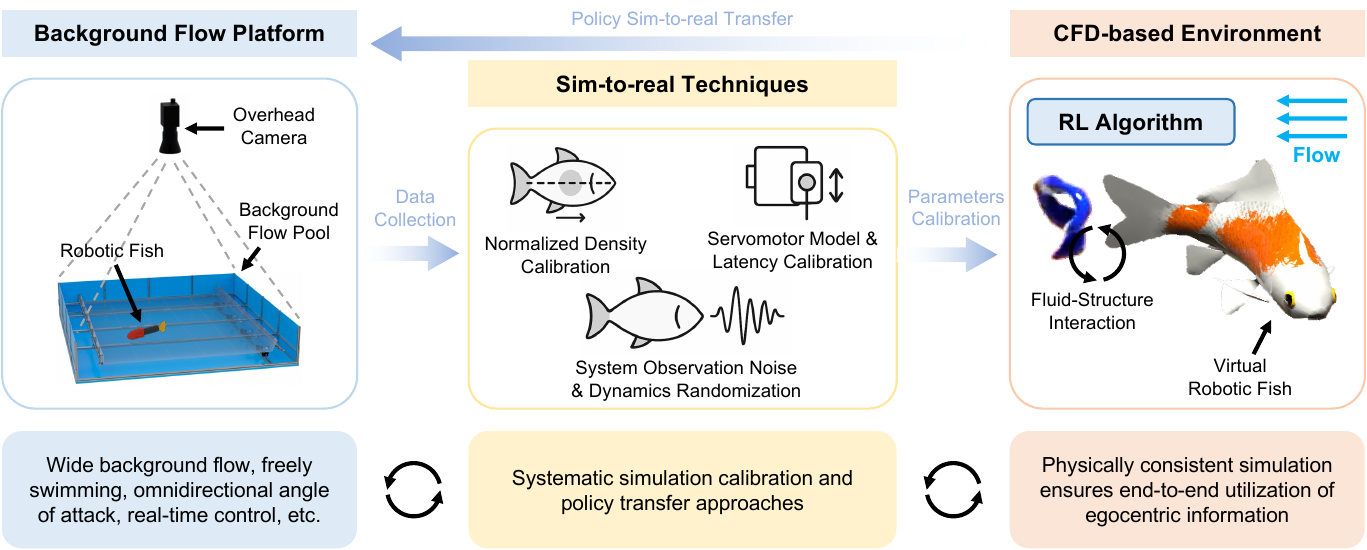}
    \caption{\textbf{Schematic diagram of the proposed framework for investigating robotic fish in unknown background flows.} (Left) Free-swimming experimental setup: a large flow tank with programmable background currents. This integrated platform enables systematic exploration of agile swimming policies for robotic fish in spatiotemporally varying flows—without explicit flow sensing. (Middle) Sim-to-real transfer module: proposed techniques achieve zero-shot policy transfer. (Right) GPU-accelerated CFD simulator: captures complex nonlinear hydrodynamics.
    }
    \label{fig:framework}
\end{figure*}

\subsection{Virtual Environment for Robotic Fish Learning}
In the past decade, RL has achieved remarkable success in addressing diverse complex robotic control tasks, such as the legged robots \cite{hwangbo2019learning,lee2020learning,miki2022learning}, flapping-wings robots \cite{kim2024wing}, and robotic fish  \cite{zhang2020path,zheng2020dual,zheng2021learning,yan2020efficient,zhang2022simulation,zhang2022leveraging}.  
Like other robotic systems, robotic fish also requires simulation-based training, as RL inherently relies on massive trial-and-error. However, their fluid-interactive characteristics render the construction of accurate simulation environments far more complex.
At present, virtual environments for robotic fish fall broadly into four categories: (i) simplified mathematical models \cite{wang2015averaging}, (ii) data-driven surrogate models \cite{yu2016data,castano2020control}, (iii) rigid-body dynamics integrated with empirical hydrodynamic formulations (e.g., sliding or added-mass models) \cite{wang2022learn}, and (iv) CFD solvers \cite{zhang2022simulation,tian2020cfd}. \textcolor{revised}{While each offers unique advantages, all suffer from critical limitations that impede their use for station-holding tasks. 
First, simplified mathematical models and their data-driven variants often overlook the complex nonlinear coupling between body deformation and unsteady fluid forces \cite{wang2015averaging,yu2016data} which prevents accurate capture of FSIs.}  
Second, while data-driven parameter fitting can partially offset unmodeled nonlinear hydrodynamics \cite{zhang2020path}, the resulting models remain confined to narrow operating regimes and require retraining as flow conditions change.
Third, although rigid-body simulators with empirical hydrodynamic terms typically deliver higher accuracy, they are constrained by oversimplified drag/lift formulations that assume quasi-steady or uniform flow. These simplifications fail to capture spatially and temporally varying flow effects that are critical to behaviors like station holding or flow harnessing.

For CFD methods, the conventional ones, including the Finite-Difference Method (FDM) \cite{ferziger2019computational,shukla2007very,strikwerda2004finite}, Finite-Volume Method (FVM) \cite{eymard2000finite,jasak2020practical,moukalled2015finite}, and Finite-Element Method (FEM) \cite{anderson1995computational,zienkiewicz2005finite}, excel at capturing detailed flow fields and complex hydrodynamic interactions.  
However, their computational cost is prohibitively high for RL training \cite{moukalled2015finite}. While industry-standard tools (e.g., ANSYS FLUENT \cite{manual2009ansys}) leverage FVM for its conservativeness, all these methods depend on body-fitted meshes, which severely restrict flexibility and efficiency, especially for unsteady FSIs \cite{manual2009ansys}. 

\textcolor{revised}{
Recently, GPU-accelerated LBM-based CFD solvers \cite{li2020fast} offer attractive computational efficiency compared to conventional solutions. However, making them reliable and efficient tools for RL training—and enabling zero-shot transfer—remains challenging due to the computational chasm between high-fidelity fluid dynamics and data-hungry learning algorithms. A data-efficient calibration and transfer pipeline to overcome the inevitable simulation-to-reality gap, is still absent.}

\section{Overview}\label{sec:overview}

\textcolor{revised}{As shown in \Cref{fig:framework}, for station-holding studies utilizing SWiFT, the workflow begins with collecting concise yet informative experimental measurements to characterize the core characteristics of the robotic fish and the surrounding flow field. These measurements are used to calibrate parameters of the CFD-based simulation environment, including those governing the robotic fish’s rigid-body dynamics and the fluid interactions. With a well-calibrated simulation environment, diverse control strategies ranging from classical controllers to RL-based policies are efficiently designed, tested, and refined in simulation. Finally, leveraging the embedded sim-to-real transfer capabilities of SWiFT, the trained strategies are seamlessly deployed to the physical robotic fish for validation under real-world flow conditions. This workflow constitutes a closed-loop development cycle that integrates data acquisition, model calibration, controller training, and real-world validation, highlighting the modularity and generalizability of SWiFT across diverse fluid-interactive robotic applications.}

\begin{figure*}[tb]
    \centering
    \includegraphics[width=0.99\linewidth]{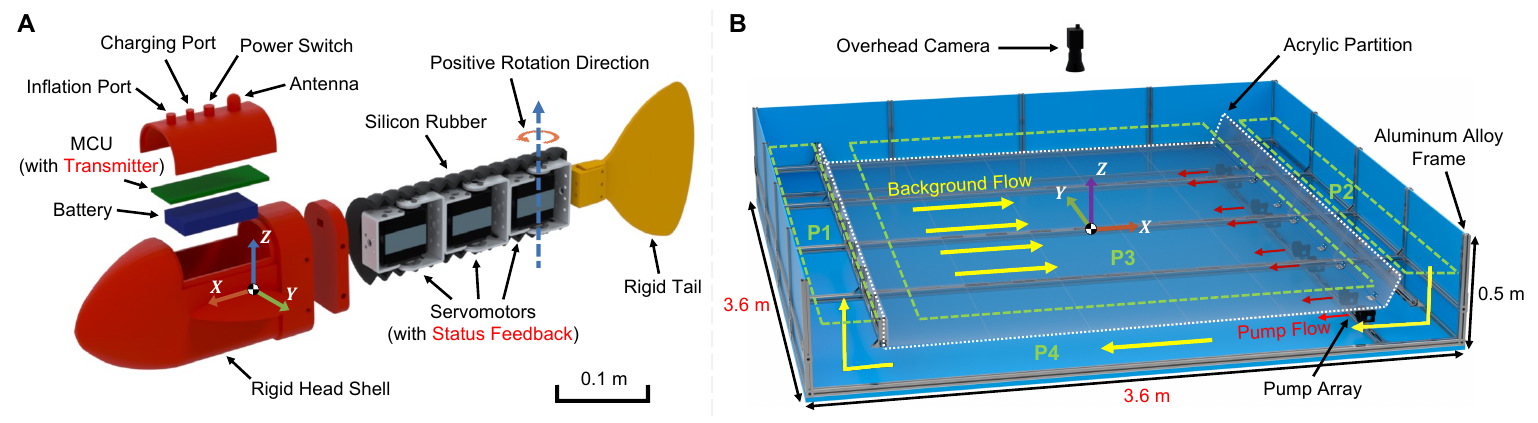}
    \vspace{-10pt}
    \caption{\textbf{The proposed experimental platform for free-swimming robotic fish in programmable background flows.} (\textbf{A}) Exploded view of the 0.52 m robotic fish prototype, highlighting key components: a rigid head shell housing the battery, MCU, charging/inflation ports, antenna, and power switch; an articulated body with three feedback-enabled servomotors encased in silicone rubber; and a rigid tail fin. (\textbf{B}) Custom 3.6 m $\times$ 3.6 m $\times$ 0.5 m flow tank, equipped with a pump array and acrylic partition to generate tunable background flows. Spanning six times the fish’s body length, this unique platform allows the free-swimming robotic fish to interact with flows (uniform currents with turbulence) within a measurement volume monitored by an overhead camera. Unlike constrained setups, it serves as a critical testbed for evaluating control strategies under more realistic hydrodynamic conditions.}
    \label{fig:platform}
\end{figure*}

\begin{figure}[tb]
    \centering
    \includegraphics[width=0.99\linewidth]{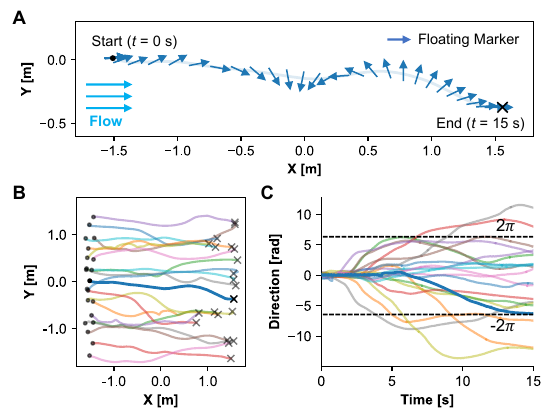}
    \vspace{-22pt}
    \caption{\color{revised}\textbf{Intuitive characterization of the complexity of the background flow tank.} (\textbf{A}) The trajectory and heading evolution of a lightweight floating marker drifting from upstream to downstream within the flow tank, capturing lateral drifts and significant rotations driven by local fluid vortices. (\textbf{B}) The drifting trajectories across 21 repeated trials. The black dots and crosses denote the start and end positions of the floating marker, respectively. (\textbf{C}) The time history curves of the heading of the floating marker across 21 repeated trials. The dashed line represents cumulative rotation exceeding one full  2$\pi$~rad turn. The results statistically demonstrate that our background flow tank successfully generates a complex and challenging turbulent background flow, has a potential consistency with natural outdoor streams.}
    \label{fig:flow}
\end{figure}

\section{Robotic Fish and Experimental Platform}\label{sec:platform}

In this section, we detail the physical experimental platform (see \cref{fig:platform}) of the SWiFT framework, which is developed to investigate the impacts of turbulent background flow on the locomotion dynamics of robotic fish, and to evaluate the station-holding control policy under background flow. The platform comprises three core subsystems: a BCF-type robotic fish featuring joint status and \textcolor{revised}{power consumption} feedback, a custom recirculating flow tank engineered to generate controllable turbulent background flow, and a host computer system integrated with an overhead RGB camera for global visual tracking and tested control policy deployment. The background flow field is predominantly laminar with superimposed low-intensity turbulence, providing a realistic hydrodynamic setting. The evaluated policy runs on the host computer and wirelessly transmits the calculated joint position commands to the robotic fish via WiFi at frequency of 50~Hz.

\subsection{Robotic Fish Prototype}

The robotic fish is developed to mimic a typical carangiform fish and adopts the BCF propulsion mode in this work (see \cref{fig:platform}A). Its overall dimensions are 0.54~m $\times$ 0.13~m $\times$ 0.12~m with a total weight of 1.1~kg, and the lengths of its four links are 0.215~m, 0.07~m, 0.07~m, and 0.18~m, respectively. The robotic fish comprises three core parts: (i) a streamlined rigid head integrating an on-board ESP32-S3 micro-controller unit, {\color{revised} a TI INA226 power measurement module} and a 7.4~V, 800~mAh lithium battery; (ii) a compliant body with three serially connected aluminum joints encapsulated in rubber skin for flexible underwater deformation; (iii) a rigid caudal fin. Both the head shell and fin are 3D-printed from ABS plastic and coated with epoxy resin for waterproofing. Each joint is actuated by a Hiwonder LX-824 bus servomotor, delivering a maximum stall torque of 17~kg·cm and a no-load speed of 0.2~s per 60°. The rubber skin is seamlessly bonded to the head and caudal fin with silicone adhesive, fully sealing internal electronics against water ingress during dynamic swimming. A key feature setting our prototype apart is its real-time joint position feedback, enhancing controller state representations. Given the focus on planar station-holding motion, the density of the robotic fish is adjusted to be slightly lower than water via counterweights, keeping it near the water surface with most of its body submerged. Its center of gravity is deliberately set low, alongside two fixed side fins on the head, to minimize roll motion induced by propulsion.

\subsection{Background Flow Tank}

We develop a recirculating flow tank to evaluate the station-holding performance of robotic fish in the turbulent background flow (see \cref{fig:platform}B). The tank comprises an aluminum alloy frame, acrylic baffles, and a pump array with eight waterproof motors, with overall dimensions of 3.6~m $\times$ 3.6~m $\times$ 0.5~m. Specifically, a horizontal acrylic partition is installed mid-depth, dividing the tank into an upper working section and a lower return channel to facilitate a large-scale recirculating loop. When the pump array beneath the partition is activated, it draws water from zone P2 into P4, establishing a water level gradient between P1 (high-level) and P2 (low-level). Driven by gravity, the fluid in P1 forms a steady overflow across the partition into P2, thereby generating a highly stable and uniform background flow field within the experimental area P3. The overall flow speed is regulated by modulating the revolutions per minute (RPM) of the pump motors, up to 0.4~m/s. \textcolor{revised}{The intuitive flow field characterization (see \cref{fig:flow}) demonstrates that the generated flow field exhibits stable and relatively uniform with low-intensity turbulence, replicate the hydrodynamic characteristics of natural streams to an extent.}

Compared with existing works that directly use motors for flow generation \cite{lin2024dynamic,xue2023experiment}, our flow tank generates a more stable, uniform, and speed-adjustable background flow field. Unlike the narrow flow tanks in \cite{zhou2010study,jung2013flow,zheng2021learning}, our upper-lower flow-guiding structure enables a larger, wider testing area with the same footprint. These advantages support unrestricted free swimming of the robotic fish in flow field, including the 360° turning and trajectory tracking task. 

\subsection{Host Computer System}

The developed host computer is paired with an overhead RGB camera, enabling real-time visual localization and data processing for deployed control algorithms. The RGB camera is mounted 3 m above the tank to capture the full water surface. The host computer detects the position and orientation of the robotic fish via its color features (\ie, red head and yellow neck) from the images. This global coordinate information is further processed into local-frame motion states, linear and angular velocity, for the deployed control algorithms. Simultaneously, the host computer receives real-time joint state feedback from the onboard MCU of the robotic fish via WiFi protocol. These data are processed by the control algorithms to generate desired joint position commands, which are wirelessly transmitted to the robotic fish, enabling unconstrained free movement in the tank. During experiments, all host computer code scripts exchange data via the robot operating system (ROS) \cite{quigley2009ros} at a frequency of 50Hz.

\begin{figure*}[t]
    \centering
    \includegraphics[width=0.99\linewidth]{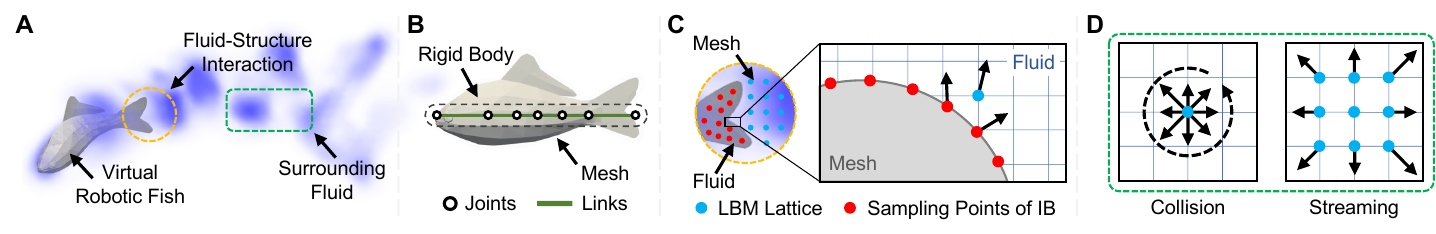}
    \vspace{-10pt}
    \caption{\textbf{Virtual environment for robotic fish.} (\textbf{A}) Overview of the simulation environment, illustrating bidirectional coupling between the robotic fish and the ambient fluid field. (\textbf{B}) The robotic fish body is modeled as an articulated rigid skeleton enclosed by a deformable mesh surface; joint actuation drives the bending of connected links, which in turn induces continuous deformation of the outer skin mesh. (\textbf{C}) The fluid field is discretized and solved via the LBM, while the body surface is sampled with IB points that mediate momentum exchange between the robotic fish and the fluid, where a parametric boundary treatment model is integrated. (\textbf{D}) The LBM solver alternates between collision and streaming steps to evolve fluid dynamics across the lattice grid. Collectively, these components form a physically consistent and computationally efficient framework that captures the hydrodynamic interactions critical for robotic fish locomotion and control.}
    \label{fig:fishgym}
\end{figure*}

\section{CFD-based Virtual Environment}\label{sec:simulation}

\subsection{Virtual Environment for Training}

To design and evaluate motion controllers for robotic fish in complex flow environments, it is critical to accurately capture the unsteady FSI dynamics around the robot, which are generally governed by the unsteady isothermal incompressible Navier-Stokes equations \cite{panton2024incompressible}:
\begin{equation} \label{eq:flow-equation}
\begin{aligned}
\nabla \cdot \mathbf{u} &= 0,\\
\rho_0\left(\frac{\partial \mathbf{u}}{\partial t} + \mathbf{u} \cdot \nabla \mathbf{u}\right) &= -\nabla p + \mu\nabla^2 \mathbf{u} + \mathbf{F},
\end{aligned}
\end{equation}
where $\mathbf{u}$ and $p$ denote the fluid velocity and pressure fields, respectively; $\rho_0$ is the constant fluid density; $\mu$ is the fluid dynamic viscosity; and $\mathbf{F}$ is the external force density.
As discussed in \cref{sec:related}, conventional CFD solvers are often cumbersome and time-intensive, while lacking the flexibility to handle dynamic FSIs (e.g., the robotic fish swimming studied here). These strict requirements render traditional CFD solvers impractical for RL-based control policy exploration, which typically demands a large volume of simulation runs. To address this challenge, we utilize a novel LBM-based FSI solver \cite{liu2022fishgym}, leveraging GPU acceleration to approximate \cref{eq:flow-equation}. Building on this solver, we employ the immersed boundary (IB) method \cite{liu2022fishgym}, integrated with our simplified parametric boundary treatment model, to simulate FSI between the robotic fish and its surrounding fluid (see \cref{fig:fishgym}). This modeled boundary-treatment not only provides a physically interpretable mechanism to adjust the boundary layer between the robotic fish and the ambient flow field, but also effectively reduces numerical stiffness. It enables stable simulation with coarser grids, thereby significantly improving computational efficiency while enabling alignment with the real physical system through the calibration of suitable model parameters.

Specifically, this parametric boundary treatment employs a linear blending between no-slip and full-slip conditions \cite{liu2023parametric}:
\begin{equation}
\mathbf{u}_w^{*} = (1 - s)\mathbf{u}_w + s\mathbf{u}_t,
\end{equation}
where $\mathbf{u}_w^{*}$ denotes the predicted wall velocity, $\mathbf{u}_w$ is the actual wall velocity, $\mathbf{u}_t$ represents the tangential velocity component of the ambient fluid, and $s \in [0,1]$ is the \textit{slip ratio} model parameter, quantifying the boundary layer thickness. A smaller slip ratio approximates the classical no-slip condition (characterized by a thick boundary layer), while a larger value approximates a thin boundary layer. During the calibration process detailed later, the slip ratio is iteratively tuned to align the simulated thrust and drag with those of the physical robotic fish, ensuring consistent hydrodynamic behavior for controller training.
\textcolor{revised}{
The proposed IB-LBM solver couples a Cartesian fluid grid with a Lagrangian representation of the solid boundary, eliminating the need for body-fitted meshing while requiring dense surface markers and repeated interpolation/forcing operations at the fluid-structure interface. This solver, coupled with LBM’s low numerical dissipation and capability to handle unsteady, turbulent flows, is well-suited for controller exploration and fine-tuning. Compared with traditional reduced-order models and many existing CFD solvers that exhibit substantially higher numerical dissipation, this low-numerical-dissipation CFD solver maintains high physical consistency even at coarse grid resolutions. When deployed as the environment for RL exploration, it enables efficient, straightforward acquisition of transferable control policies with limited dedicated measures to calibrate, and achieve effective control on physical robotic fish. Moreover, the inherently local and explicit properties of LBM are highly amenable to GPU acceleration, yielding significant improvements in efficiency.}

Nevertheless, the GPU acceleration alone, while necessary, is insufficient to support millions of interaction steps required for RL.  
A deliberate trade-off between simulation accuracy and computational efficiency is required, primarily dictated by the spatial and temporal resolution of the simulator. Higher resolutions capture finer turbulent structures and detailed flow-body interactions but lead to drastically increased computational cost, which renders RL impractical within a reasonable timeframe.  
In our case, the Reynolds number of the robotic fish places it in a regime where mean flow effects dominate the hydrodynamic forces on its body, rather than fine-scale turbulent fluctuations. 
We thus use a moderate spatial and temporal resolution assisted by a turbulence model \cite{nicoud1999subgrid} to boost efficiency, while retaining the essential flow behavior critical for training control policies, achieving simulation of 3 seconds of physical time in just 1 second of computation on a standard personal computer (see \cref{sec:results} for specific simulator setting in this work). 
However, this coarsening inevitably introduces modeling inaccuracies and uncertainties. To ensure policies trained under these conditions remain transferable to the physical robot, we propose a set of simulation calibration techniques, detailed in \cref{sec:simulation}-C.

\subsection{Virtual Robotic Fish Setting}

For solid geometries with fine surface details (e.g., skin wrinkles or micro-textures), dense sampling is required for IB-LBM, which increases computational cost and potentially introduces numerical stiffness due to mismatches between grid resolution and sample density. To mitigate the impact of these issues, we simplify the geometric model of the robotic fish under the premise that detailed shape features have minimal influence on RL training.
As illustrated in \cref{fig:fishgym}, small-scale surface wrinkles and high-frequency details are smoothed, while the overall body shape, wetted surface area, tail geometry, and joint kinematics are preserved. 
This streamlined morphology suppresses spurious oscillations at the fluid-structure interface and enhances the stability of long-horizon simulations. The resulting simplified geometry takes the form of a koi-inspired robotic fish model, our virtual robotic fish for this study.
However, this geometric simplification inevitably introduces morphological and actuation discrepancies compared to the physical robot detailed in \cref{sec:platform}, which modifies flow structures and may impact the control policy. Consequently, the learned policy must address not only the general sim-to-real hydrodynamic gap but also residual inaccuracies in body shape and actuator dynamics, which motivates the simulator calibration techniques detailed in the \cref{sec:simulation}-C.

\begin{figure*}[t]
    \centering
    \includegraphics[width=0.99\linewidth]{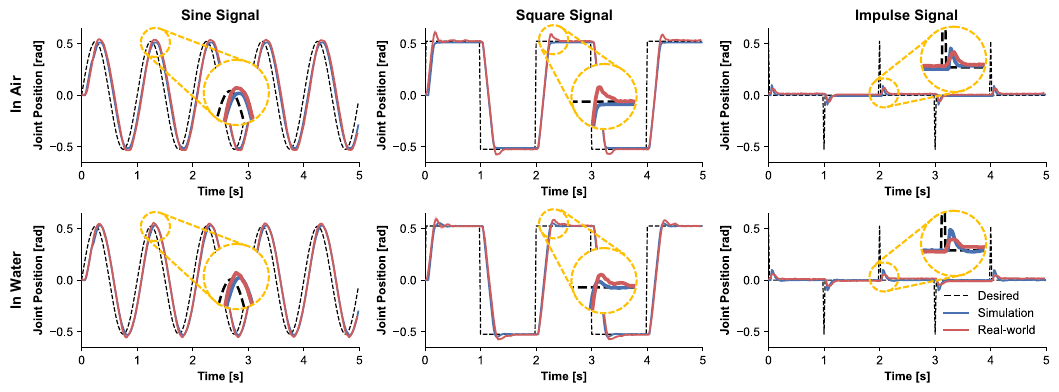}
    \vspace{-10pt}
    \caption{\color{revised}\textbf{Dynamic response verification between the simulation and the real world.} The time history curves of the desired and the actual joint position generated by three representative command signals: sine (left), square (middle), and impulse (right), in both simulation and the real world. The evaluation conducted in both air and water mediums to validate load fluctuations of the servomotor. The robotic fish was suspended to avoid any mechanical interference from clamps or fixtures on the root joint dynamics in the air scenario. The high quality consistency among the desired trajectories (dashed black), simulation responses (blue), and measured real-world responses (red) verified that the effectiveness of the proposed calibration techniques.}
    \label{fig:calibration}
\end{figure*}

\subsection{System Synthesis and Calibration}

\textcolor{revised}{
The goal of calibration is to ensure that the simulator reproduces the comparable behaviors of the physical robotic fish under same inputs.
By matching the measurable input–output behavior of each subsystem, the calibrated simulator becomes a dynamically consistent surrogate of the hardware, enabling reliable policy learning and transfer.
}

\subsubsection{Actuator Model Calibration}

While our simulator offers a torque-controlled environment for FSI, the physical robotic fish prototype uses position-controlled servomotors.
This fundamental actuation interface mismatch prevents directly trained simulator policies from being deployed on the physical system. Additionally, direct torque measurement or control is impractical for the hardware, making actuator-level calibration critical to align control dynamics between simulation and experiment.
To bridge this actuation gap, we implement a first-order proportional-derivative (PD) controller in simulation to emulate servo dynamics:  
\begin{equation}
\mathbf{T} = \mathbf{k_p} (\mathbf{J}^\text{des} - \mathbf{J}) - \mathbf{k_d} \dot{\mathbf{J}},
\end{equation}  
where $\mathbf{J}^\text{des}$ and $\mathbf{J}$ are the desired and actual joint angles, and $\mathbf{T}$ is the torque input to the simulator. The proportional ($\mathbf{k_p}$) and derivative ($\mathbf{k_d}$) gains, both $\in \mathbb{R}^3$, are tuned to match the simulated joint dynamics to the transient response of the physical servomotors under identical excitation.

\subsubsection{Communication Latency Modeling}

Wireless communication between the host computer and the robotic fish introduces non-negligible latency.  
To replicate this effect, we keep a temporal buffer of past commands $\mathbf{J}^\text{des}_{t}$ and apply a delayed command based on the measured latency $\Delta t$:  
\begin{equation}
\mathbf{J}^\text{used}_t = \mathbf{J}^\text{des}_{t - \Delta t},
\end{equation}  
which aligns the control timing in simulation with that of the real system, enhancing the fidelity of control-loop dynamics.

\subsubsection{Friction Calibration}

Hydrodynamic friction between the body of the robotic fish and the surrounding fluid environment cannot be perfectly modeled, owing to complex boundary-layer effects.  
To achieve hydrodynamic consistency between simulation and experiment, we conduct an open-loop characterization of the robotic fish under identical actuation inputs. 
Specifically, each joint is driven by a bias-free sinusoidal signal with a frequency of 1~Hz and an amplitude of 30°. 
During the motion, the translational velocities along the $x$- and $y$-axes as well as the angular velocity about the yaw axis are recorded over time for both the real and simulated systems. 
The previously introduced slip-ratio coefficient is then iteratively tuned to minimize the discrepancy between these velocity profiles, such that the simulated responses faithfully reproduce the measured hydrodynamic behavior of the physical robot. 
This procedure effectively aligns the resistive and propulsive characteristics across the two domains, establishing a consistent frictional interaction model for subsequent closed-loop control and learning experiments.

\subsubsection{Mass Distribution Calibration}

An accurate mass distribution is key to replicating the robot’s inertial properties and rotational dynamics.  
We fine-tune the simulated model’s internal mass parameters to match the physical robot’s measured center of mass and inertia tensor.  
This calibration reduces body-attitude response discrepancies and ensures dynamic consistency between simulation and hardware.

\vspace{0.2cm}

Together, these calibrations align the simulated actuation, hydrodynamic, and inertial characteristics with the ones of the real robotic fish, enabling the learned control policy to reliably transfer from simulator to physical system. \textcolor{revised}{We have added a set of comparative experiments between the measured response curves of physical servomotors and the corresponding simulation curves, using three representative signals: sine, square, and impulse, conducted in both air and water, with 3 times replicated for each real-world scenario, as shown in \cref{fig:calibration}. Specifically, in air scenario, the robotic fish was suspended to avoid any mechanical interference from clamps or fixtures on the root joint dynamics. The same three signals were then applied in simulation, both with and without fluid interaction function enabled. The comparative results reveal a high degree of consistency between the physical and simulated response curves in both air and water, which verified the high fidelity of the servomotor model implemented in the simulation. }

\section{RL-based Training}\label{sec:method}

In this section, we detail the development of a robust station-holding policy for robotic fish via our SWiFT system. The policy is purely \textcolor{revised}{egocentric}, relying solely on local coordinate information to counteract background flow, without prior knowledge of the flow field (e.g., direction or magnitude) or any flow perception sensor.

\subsection{Reinforcement Learning}

We formulate the station holding of robotic fish in unknown background flow as a Markov decision process (MDP), defined by the tuple $(S, A, R, D, P_{sas'})$. Here, $S$ is the state space, $A$ the action space, $R$ the reward function, $D$ the initial state $s_0$ distribution, and $P_{sas'}$ the state transition probability.
At the start of each episode, the environment samples $s_0$ from $D$. The agent selects the optimal action $a_0$ via policy $\pi$ from $s_0$, after which the environment returns a reward $r_0$ and transitions to the next state $s_1$ per $P_{sas'}$. This process repeats until termination conditions end the episode.

We solve the above MDP for the virtual robotic fish in simulation using RL, learning an optimal policy $\pi^{*}$ that maximizes the episode’s cumulative reward (expected return):  
\begin{equation}
\pi^{*} = \arg\max_{\pi} E_{\tau(\pi)}[\sum_{t=0}^{T} \gamma^{t}r_t],
\end{equation}  
where $\gamma \in [0, 1)$ is the discount factor (ensuring a finite expected return) and $T$ is each episode’s horizon. Details of the MDP’s elements are as follows:

\subsubsection{State Space}

To preserve the potentiality to deployment to real-world aquatic settings in the future, we deliberately use state representations in the local robot coordinate system (i.e., egocentric perception) rather than relying on inertial-frame representations (e.g., global water tank coordinates) as in \cite{wang2024investigation,zhou2010study,jung2013flow}.  
To this end, we define the state as:  
\begin{equation}
\mathbf{s}_t = [\mathbf{q}, \mathbf{v}, w, \mathbf{J}, \mathbf{K}],
\end{equation}  
where $\mathbf{q} \in \mathbb{R}^2$ (target position vector), $\mathbf{v} \in \mathbb{R}^2$ (linear velocity vector), and $w \in \mathbb{R}$ (angular velocity) of the robotic fish  are all represented in the robot’s centroid coordinate system. $\mathbf{J} \in \mathbb{R}^3$ (joint angular positions) and $\mathbf{K} \in \mathbb{R}^3$ (joint angular velocities) are in radians, defined in the robot’s joint coordinate system.  
The centroid coordinate system has its $x$-axis pointing forward and $y$-axis pointing left shown in \cref{fig:platform}A.

Due to the highly nonlinear, spatiotemporally varying nature of background flows, sensing, characterizing, and utilizing flow-field data usually require additional onboard sensors and sophisticated data processing algorithms \cite{zheng2021learning, panta2025leader, jung2013flow}. 
Inspired by biological studies \cite{liao2006role}, we instead explore the limits of the \textcolor{revised}{egocentric} capability of the robotic fish, deliberately excluding flow-related data from the state definition and only augmenting the state space with joint-related information $\mathbf{J}$ and $\mathbf{K}$.  
The joint angular position $\mathbf{J}$ largely characterizes the shape of the robotic fish, helping the controller establish the relationship between body shape and background flow-induced additional forces and enabling appropriate bending/swinging actions across different states (see \cref{fig:saliency}).

\subsubsection{Action Space}

For motion control of robotic fish, most existing methods use CPG, with CPG bias parameters as the action space. While effective in still water, CPG-based approach significantly limits agility \cite{lin2025learning}, a critical flaw for station holding in flowing environments, as it demands high agility, multi-modal motion, and environmental resilience.
To address this, we directly define the action space as each joint’s desired angular positions: $\mathbf{a}_{t} = [\mathbf{J}^D] \in \mathbb{R}^3$. This end-to-end policy eliminates the need for a CPG module.
Unlike prior works that rely on CPGs for indirect joint control \cite{yan2020efficient, zhang2022simulation, zhang2022leveraging}, our method is a proven alternative \cite{lin2025learning}. It not only avoids the tedious parameter tuning of CPG controllers but also delivers superior agility and multi-modal motion capabilities---key factors for successful station holding.

\subsubsection{Reward Function}

For the station-holding task, the robotic fish must not only approach the target position quickly and remain there without overshoot but also move its joints smoothly throughout the entire process. The reward $r_t$ at each step $t$ has two main components:  
\begin{equation}
r_t = w^G r^G_t + w^S r^S_t,
\end{equation}  
where $r^G_t$ and $r^S_t$ are the rewards for station holding and smooth joint motion, and $w^G, w^S$ are their corresponding weights.  
Specifically, the station-holding objective is:  
\begin{equation}
r^G_t = (1 + w^D d)^{-1},
\end{equation}  
where $d$ is the straight-line distance between current position of the robotic fish and the target. This design ensures that $r^G_t$ increases rapidly as $d$ decreases, encouraging the robot to quickly approach the target and maintain high positional accuracy near it.  
The smooth joint motion objective is:  
\begin{equation}
r^S_t = -||\mathbf{K}||_{2}^{2},
\end{equation}  
where $\mathbf{K} \in \mathbb{R}^3$ is the vector of each angular velocity of the joint. By constraining sudden changes in joint angular velocity, this objective minimizes twitching from discontinuous motions, enhancing movement smoothness and significantly reducing swimming energy consumption.
Notably, joint twitching also hinders the sim-to-real transfer of the trained policy. 
For example, Xue et al. \cite{peng2018sim} mitigate twitching behavior by adding a low-pass filter to action outputs, whereas our approach targets the root cause and enhances sim-to-real transfer by directly constraining joint angular velocity.

\begin{figure*}[tb]
    \centering
    \includegraphics[width=0.99\linewidth]{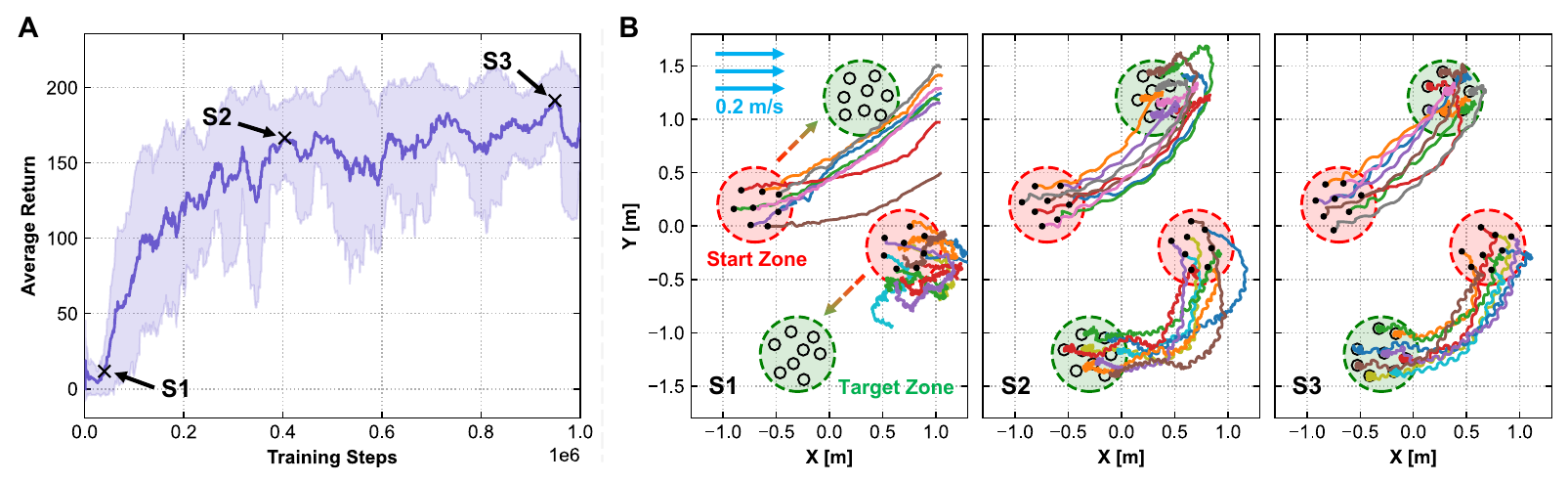}
    \vspace{-10pt}
    \caption{\color{revised}\textbf{Training process and representative stages evaluation of the station-holding policy.} (\textbf{A}) The average return over training steps demonstrates robust, efficient policy convergence, achieved 10\textsuperscript{6} steps (approximately 16 hours) of training, with results aggregated as mean ± standard deviation across three independent runs. (\textbf{B}) Evaluation trajectories at three representative training stages in simulation, capturing the behavioral evolution from uncoordinated motion (S1) and basic target approaching (S2) to robust station holding (S3). All evaluations are conducted within a 0.2~m/s background flow across eight random trials for both right-rear downstream and left-front upstream scenarios.}
    \label{fig:training_process}
\end{figure*}

\subsubsection{Initial State Distribution}

To develop a generalizable station-holding policy, we uniformly sample the initial positions of the robotic fish and targets within a [-1.5, 1.5] meter square at the start of each training episode. The robot’s head orientation is also uniformly sampled over [-$\pi$, $\pi$] radians.
Background flow velocity is uniformly sampled within [0.1, 0.3]~m/s, and flow direction is uniformly sampled over  [-$\pi$, $\pi$] radians.
These initial state distributions expose the policy to diverse target positions and flow conditions during training, enhancing the feasibility and robustness of the robot across scenarios. Notably, this random sampling includes scenarios where the robotic fish faces away from the incoming flow— a condition absent in prior work \cite{jung2013flow, salumae2013flow}. This avoids the limitation of policies restricted to specific scenarios or requiring retraining for different flow conditions.

\subsubsection{Termination Condition}

During training, episodes terminate only when the robotic fish swims out of the simulation domain or reaches the maximum steps $T_{max} = 150$ of current episode. The control interval is 0.2 s. The training environment then restarts a new episode by randomly sampling from the initial state distribution.
It is worth mentioning that most existing works train robotic fish swimming policies with position arrival as the objective, terminating episodes immediately upon target reach. However, such policies fail to teach the robotic fish to maintain position or resist background flows, leading to significant overshoot or outright failure when applied to our station-holding task.
In contrast, our station-holding training eliminates the termination-upon-arrival condition. It combines position control with holding requirements integrated into the reward function and termination logic, forcing the robotic fish to learn the nonlinear dynamics of both itself and the surrounding fluid. This enables the robot to leverage lift and drag from fluid interactions to counteract overshoot and achieve precise station-holding control.

\subsection{Dynamics Randomization and Observation Noise}

\textcolor{revised}{To enable zero-shot transfer of the trained policy to the physical robotic fish, we supplement the calibration techniques in \cref{sec:simulation} with dynamics randomization \cite{peng2018sim} and observation noise, varying key system parameters during training to enhance the policy’s robustness to real-world dynamics. We uniformly sample the mass of each robotic fish link $\pm$10\% around its original value, the servomotor model’s response parameters $K_p$ and $K_d$ $\pm$10\% and $\pm$5\% around their original values respectively, and the joint mechanical zero positions within [-5°, 5°] in simulation, accounting for physical servomotor assembly errors. These samplings occur at the start of each training episode and keep constant until the episode ends. Additionally, to address observation errors and sensor noise, the robotic fish’s states are augmented with unbiased Gaussian noise matching the physical robotic fish, improving the policy’s robustness to system uncertainty.}

\begin{figure*}[tb]
    \centering
    \includegraphics[width=0.99\linewidth]{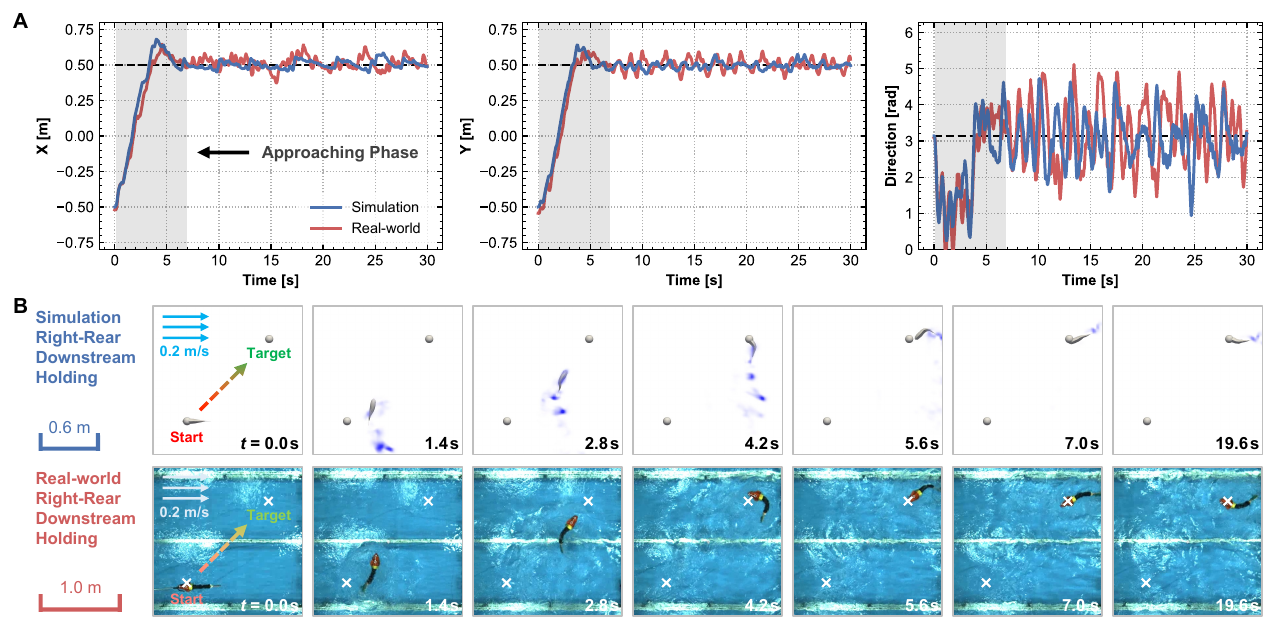}
    \vspace{-10pt}
    \caption{\color{revised}\textbf{Sim-to-real comparison of right-rear downstream station-holding performance.} (\textbf{A}) The position and heading curves of the controlled robotic fish during right-rear downstream station holding in 0.2~m/s background flow in both simulation and real-world experiments. (\textbf{B}) Sequential snapshots of the approaching (0--7~s) and holding (7--30~s) phases during right-rear downstream station holding in both simulation and real-world experiments. The high consistency curves and snapshots indicates that the trained policy achieved sim-to-real transfer, and the controlled robotic fish successfully achieved rapid target approaching, robust station holding, and spontaneous rheotaxis behavior.} 
    \label{fig:sim2real_downstream}
\end{figure*}

\begin{figure*}[tb]
    \centering
    \includegraphics[width=0.99\linewidth]{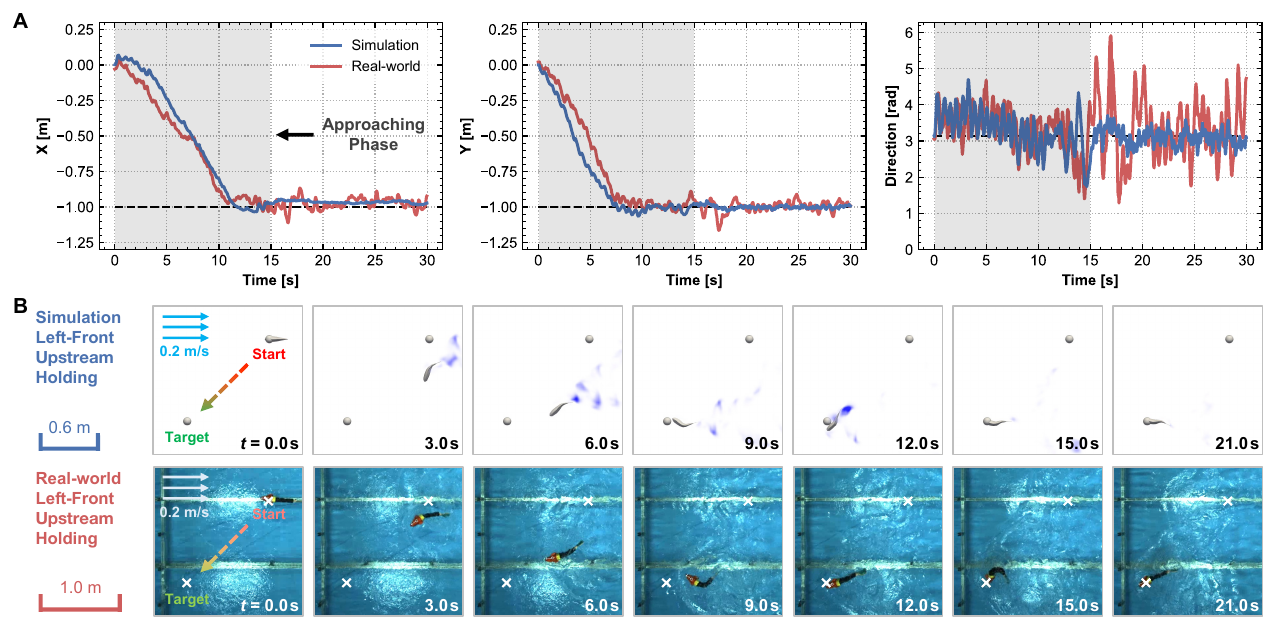}
    \vspace{-10pt}
    \caption{\color{revised}\textbf{Sim-to-real comparison of left-front upstream station-holding performance.} (\textbf{A}) The position and heading curves of the controlled robotic fish during left-front upstream station holding in 0.2~m/s background flow in both simulation and real-world experiments. (\textbf{B}) Sequential snapshots of the approaching (0--15~s) and holding (15--30~s) phases during left-front upstream station holding in both simulation and real-world experiments. The high consistency curves and snapshots once again fully indicates that the trained policy achieved sim-to-real transfer, and the controlled robotic fish successfully achieved rapid target approaching, robust station holding, and spontaneous rheotaxis behavior.} 
    \label{fig:sim2real_upstream}
\end{figure*}

\begin{figure*}[tb]
    \centering
    \includegraphics[width=0.99\linewidth]{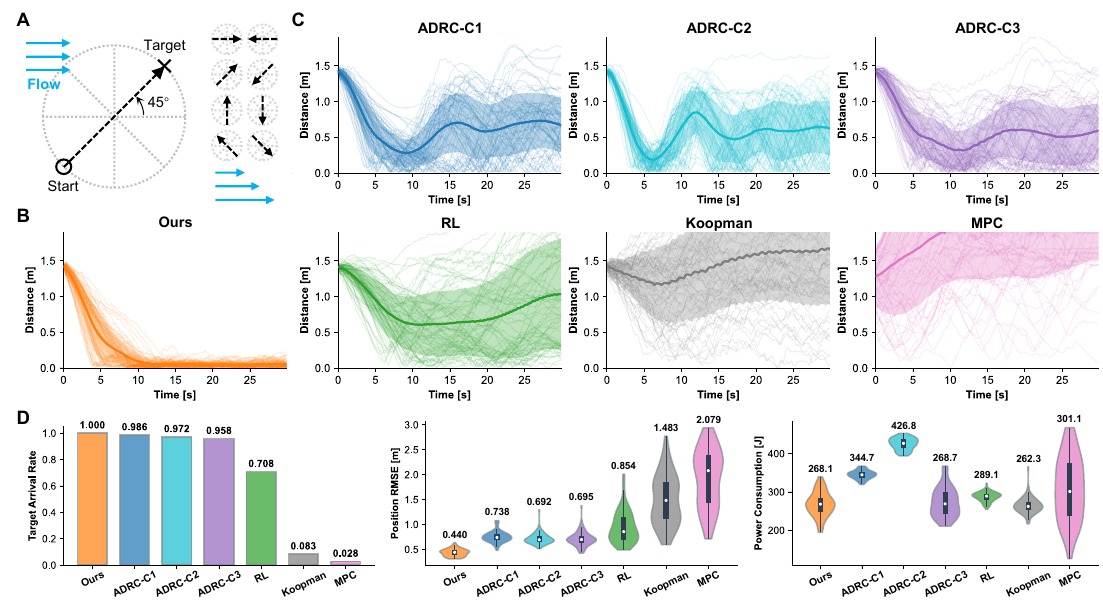}
    \vspace{-10pt}
    \caption{\color{revised}\textbf{Comprehensive comparison of station-holding performance under different control policies.} (\textbf{A}) Schematic of the experimental protocol, encompassing 3 background flow speeds (0.1, 0.2, 0.3~m/s), 8 initial target line-of-sight (LOS) angles (0° to 315° with 45° intervals), and 3 independent trials per condition, yielding a total of 72 trials for each control policy. (\textbf{B})--(\textbf{C}) Distance curves across all 72 trials for the proposed control policy (\textit{Ours}), three different parameter configurations of ADRC controller \cite{li2019bottom} (\textit{ADRC-C1}, \textit{ADRC-C2}, \textit{ADRC-C3}), alternative CFD-tuned RL control policy \cite{zhang2022simulation} (\textit{RL}), Koopman-based controller \cite{mamakoukas2021derivative} (\textit{Koopman}), and MPC controller \cite{wang2025identification} (\textit{MPC}). The thin lines, bold lines, and shaded areas denote all trajectories, the average trajectories, and distribution variances, respectively. (\textbf{D}) Quantitative statistical evaluation across all 72 trials per policy based on three metrics: target arrival rate (left), position root-mean-square error (RMSE) (middle), and power consumption (right).}
    \label{fig:comparison_sota}
\end{figure*}

\begin{figure*}[tb]
    \centering
    \includegraphics[width=0.99\linewidth]{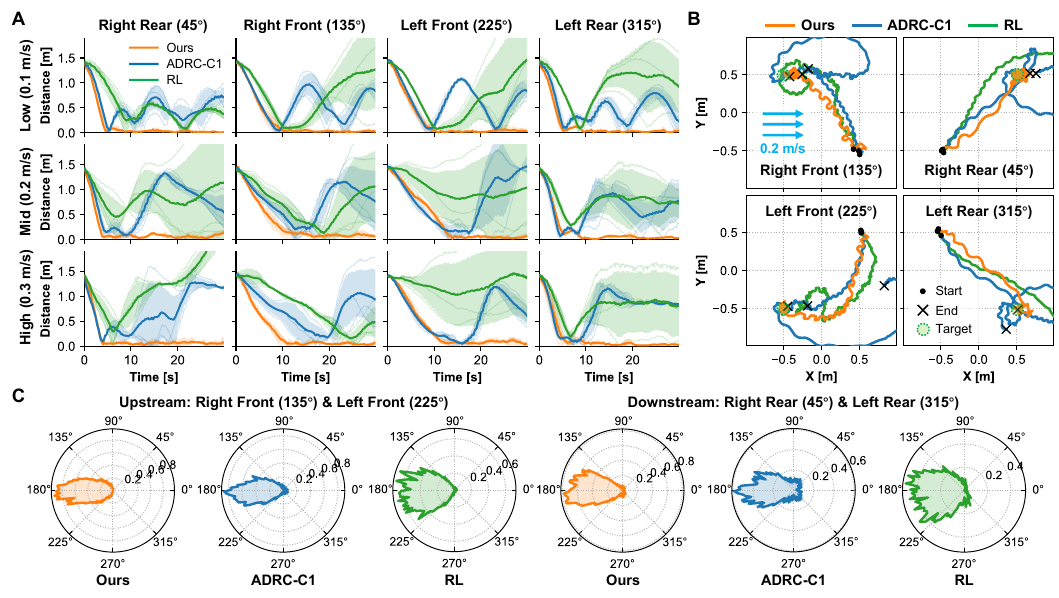}
    \vspace{-10pt}
    \caption{\color{revised}\textbf{Comparative analysis of station-holding performance under representative control policies.} The evaluation compares the proposed control policy (Ours) against the ADRC-C1 baseline \cite{li2019bottom} and the alternative RL baseline \cite{zhang2022simulation} under various turbulent background flow conditions. (\textbf{A}) Comparative distance errors over time across all flow speeds (0.1, 0.2, and 0.3~m/s) and representative initial target line-of-sight (LOS) directions (45°, 135°, 225°, and 315°), highlighting the fast convergence and minimal steady-state error of \textit{Ours}. The thin lines, bold lines, and shaded areas denote all trajectories, the average trajectories, and distribution variances, respectively. (\textbf{B}) The trajectories of the robotic fish under a 0.2~m/s flow speed, demonstrating that our trained policy takes a more direct path and eliminates severe target overshoots or wide loitering. (\textbf{C}) The polar performance profiles quantifying the attitude variance under upstream (right front \& left front) and downstream (rear right \& left rear) scenarios, substantiating the tight, directionally uniform accuracy achieved by our framework compared to the highly variable errors of the \textit{ADRC-C1} and the \textit{RL}.}
    \label{fig:comparison_qualitative}
\end{figure*}

\begin{figure*}[tb]
    \centering
    \includegraphics[width=0.99\linewidth]{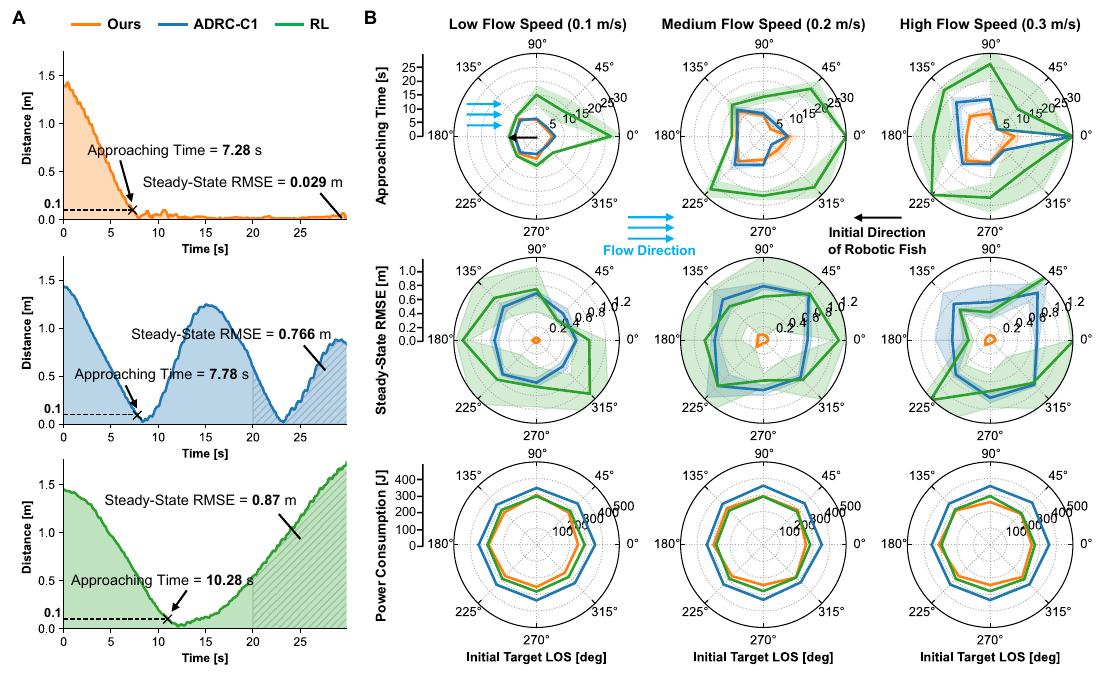}
    \vspace{-10pt}
    \caption{\color{revised}\textbf{Comprehensive analysis of station-holding performance under multi-condition coupling.}  The evaluation isolates the control responses across three distinct metrics to thoroughly elucidate the decoupling effects of varying flow velocities and initial target LOS directions. (\textbf{A}) Representative distance-to-target time-history curves illustrating the calculation of approaching time (the duration to first enter the target vicinity) and the steady-state root-mean-square error (RMSE) for the proposed method (Ours), ADRC-C1 \cite{li2019bottom}, and RL \cite{zhang2022simulation} controllers. (\textbf{B}) Radar plots summarizing the averaged evaluation metrics across eight initial target line-of-sight (LOS) directions (0° to 315° with 45° intervals) and three characteristic flow speeds (0.1, 0.2, and 0.3~m/s). The rows from top to bottom systematically contrast the approaching time (top), position RMSE (middle), and power consumption (bottom), demonstrating the superior directional uniformity, fast convergence, and balanced energy efficiency of the proposed framework under complex background currents.}
    \label{fig:comparison_quantitative}
\end{figure*}

\section{Experimental Results}\label{sec:results}

In this section, we perform extensive evaluations on the trained station-holding policy to validate the effectiveness of the proposed learning framework and our SWiFT system, highlighting the capability of our platform to achieve station-holding of robotic fish that are challenging with existing platforms and methodologies. 
First, we present the policy’s training process and then assess its sim-to-real transferability. 
Next, we compare the trained policy with SOTA algorithms to demonstrate its superiority. 
\textcolor{revised}{We also evaluate the effect of different initial angle of attack (AoA) and the biological consistent with real fish rheotaxis. }
\textcolor{revised}{We also transfer the learned station-holding policy to trajectory tracking in unknown background flows. }
Finally, we conduct ablation analyses on factors influencing training and sim-to-real transfer. 

\subsection{Training Process and Evaluation}

Policy training and evaluation in this work are performed on personal computers with an Intel Core i9-12900K CPU and an NVIDIA GeForce RTX 4090 GPU. Simulation timestep is set to $0.004$ s, domain size $4$ m$\times4$ m$\times1$ m (length $\times$ width $\times$ height), and grid resolution $200\times200\times50$.  
The actor network is a multi-layer perceptron (MLP) with the structure [11 (ReLU), 256 (ReLU), 256 (ReLU), 3 (Tanh)], and the critic network is an MLP with [14 (ReLU), 256 (ReLU), 256 (ReLU), 1].
The training process for the station-holding policy, implemented via the proposed framework, is shown in \cref{fig:training_process}A. It can be seen that the average return increases with training steps, converging at approximate $4\times10^5$ steps.

Notably, with our simulator, the full $10^6$-step training took only around 16 hours, with the pre-convergence phase ($4\times10^5$ steps) taking around 6 hours. In contrast, the current SOTA robotic fish learning framework requires around 10 days to fine-tune just 50 episodes, and an estimated 400 days to train a complete swimming policy from scratch \cite{zhang2022simulation}. This further highlights our approach’s efficiency advantage: over a 600-fold acceleration versus SOTA, especially given that station holding under background flow is far more challenging than the steady still-water swimming focused on in prior works.

To intuitively illustrate the policy’s training process, \cref{fig:training_process}B shows the station-holding performance across three key learning stages. 
For each stage, trajectories from 8 sampled trials are presented for two scenarios (downstream/upstream) to illustrate training progress, with background flow velocity set to 0.2~m/s (left to right).
In the initial stage S1, the robotic fish barely swings its tail toward the target and shows limited flow compensation: downstream initialization leads to target overshoot from background flow force, while upstream initialization hinders target approach against the current. By stage S2, the agent can overcome flow effects, achieving and maintaining the target position in both scenarios. At final convergence (S3), the virtual robotic fish exhibits agile \emph{rheotaxis} and a highly stable station-holding policy, efficiently reaching and maintaining the target position regardless of initial conditions.
These trajectories qualitatively trace the evolution of the policy, from uncoordinated motion to emerging stability and refined station holding, clearly demonstrating progressive improvement via training.

\subsection{Zero-Shot Sim-to-Real Transfer}

The trained \textcolor{revised}{egocentric} station-holding policy is zero-shot transferred to the physical robotic fish (no fine-tuning). The policy is evaluated under simulation-matching experimental conditions, tackling station-holding tasks in both downstream and upstream scenarios.  
Notably, real-world background flow contains more turbulence than in simulation due to the difficulty (and impracticality) of generating perfectly uniform flow, and since mild turbulence better mirrors realistic outdoor aquatic environments. This setup also serves as a more rigorous testbed for assessing policy robustness.  
Target positions are placed $1.4$~m diagonally ahead (left-front, upstream) and behind (right-rear, downstream) of the initial position. 

As shown in \cref{fig:sim2real_downstream}, the robotic fish reached the target position in $\leq5$~s under downstream conditions, both in simulation and real-world experiments. 
In contrast, the upstream task (\cref{fig:sim2real_upstream}) took around $10$~s, twice the downstream duration, clearly showing background flow’s influence. 
Notably, without imposed orientation targets, the robotic fish learned to first steer toward the target and then reorient upstream to hold position against the flow, a consistent behavior in downstream trials.
Comparing simulated and real-world trajectories, the policy shows high sim-to-real consistency despite large discrepancies in geometry, computing inaccuracies, and uncertainties in turbulence modeling. Video snapshots further confirm strong consistency across motions. While real-world trajectories exhibit slightly higher sensor noise, the policy performs robustly without fine-tuning, validating the effectiveness of sim-to-real transfer under realistic flow conditions.

{\color{revised}
\subsection{Comparison with SOTA Controllers}
}

\textcolor{revised}{
To validate our proposed algorithm, we extensively compare it against a diverse set of state-of-the-art (SOTA) aquatic robot controllers recently published in top-tier robotics journals. These include: (i) an established traditional benchmark \cite{li2019bottom} integrating a proportional navigation law (PNL) \cite{guelman2007qualitative} with an active disturbance rejection controller (ADRC) \cite{han2009pid}, evaluated across three parameter configurations (ADRC-C1, C2, and C3); (ii) a model predictive control (MPC) approach; (iii) a Koopman-operator-based data-driven controller; and (iv) an alternative deep reinforcement learning (RL) baseline utilizing conventional reward formulations \cite{zhang2022simulation}. 
}

{\color{revised}
Real-world comparative evaluations are comprehensively performed via station-holding tasks yielding a benchmark dataset across eight distinct initial target line-of-sight (LOS) directions and three turbulent flow velocities ($0.1$~m/s, $0.2$~m/s, and $0.3$~m/s), with the ambient flow characteristics entirely unknown to the robots. For each control policy, a total of 72 experimental trials are conducted to ensure statistical rigor.
}

{\color{revised}
As illustrated in \cref{fig:comparison_sota}, our proposed controller exhibits remarkable convergence consistency, where the distance-to-target trajectories rapidly damp out within $10$~s without noticeable overshoot across all trials. In contrast, the model-based approaches (MPC and Koopman) exhibit severe drift and diverging trajectories due to model inaccuracies stemming from highly nonlinear fluid-structure interactions. Quantitatively, our method, alongside ADRC-C1 ($98.6\%$), ADRC-C2 ($97.2\%$) and ADRC-C3 ($95.8\%$), achieves a flawless target arrival rate ($100\%$), significantly outperforming the standard RL ($70.8\%$), Koopman ($8.3\%$), and MPC ($2.8\%$). However, as highlighted by the comprehensive metric distributions, this success in ADRC variants comes at a prohibitive energetic and accuracy cost: our controller yields the lowest and most concentrated position RMSE, while reducing cumulative power consumption.
}

\textcolor{revised}{
A two-sided Welch's t-test confirms that Ours statistically outperforms the alternative controllers with overwhelming significance (p\textless 0.001 across all baseline pairs) in position RMSE, demonstrating robust station-holding capabilities under unknown turbulent background flow. Moreover, the Ours has similar power consumption with ADRC-C3 (p=0.37) and Koopman (p=0.39) policy, but Ours still outperforms the other controllers with overwhelming significance (p\textless 0.001).
}

\textcolor{revised}{
To thoroughly elucidate the decoupling effects of varying flow velocities and initial target directions, a detailed transient comparison is conducted among our framework, the standard ADRC-C1 \cite{li2019bottom}, and the alternative RL baseline \cite{zhang2022simulation}. While these three methods reach the target vicinity in similar time, the ADRC-C1 controller exhibits notably larger steady-state errors during the holding phase, while the alternative RL baseline suffers from severe stochastic oscillations. In contrast, our proposed method takes a more direct path and maintains a far tighter trajectory near the target. Notably, in the $0.2$~m/s downstream station-holding task, the ADRC trajectory shows a distinct target overshoot and undergoes wide loitering due to lower control agility, whereas the RL baseline drifts away due to slow turns. 
}

\textcolor{revised}{
To further quantify this tri-strategy comparison, we split the robot’s response into two phases: the approaching stage (the first entry into a $0.1$~m radius of the target) and the steady stage (remaining duration). Approaching time is the duration to reach the target vicinity, while steady-state performance is assessed via the total process distance root-mean-square error (RMSE), as shown in \cref{fig:comparison_quantitative}A. Results show that our proposed method achieves a steady-state RMSE of only around $0.05$~m, which is substantially lower than the $0.8$~m of the ADRC-C1 method and the $1.0$~m of the alternative RL baseline, while maintaining highly comparable power consumption.
}

\textcolor{revised}{
For a comprehensive, intuitive comparison of these three strategies, we present the averaged results as heatmaps in \cref{fig:comparison_quantitative}B. In terms of approaching time, all strategies follow a clear trend: higher flow velocity or a target direction closer to 0° (directly upstream) increases duration. This is mainly due to stronger upstream flows exerting greater hydrodynamic resistance, requiring the robot to generate higher thrust for counteraction. Meanwhile, continuous heading adjustments under strong currents further delay convergence, reflecting the inherent stability-maneuverability trade-off in flow-adaptive swimming control. Notably, our proposed algorithm achieves consistently shorter approaching times than both the ADRC and RL baselines across most conditions. More notably, our method’s steady-state distance error remains consistently low ($\approx 0.05$~m) and highly uniform across directions, which is substantially better and more stable than the larger, more variable errors exhibited by the baseline controllers.
}

{\color{revised}
\subsection{Different Initial Angle of Attack}
To evaluate the robustness and symmetry of the learned station-holding policy against varying incoming flow directions, we conduct experiments under four representative initial angles of attack (AoA): -180°, -90°, 0°, and 90°, as illustrated in \cref{fig:aoa}. 

As shown in \cref{fig:aoa}A, the policy drives the distance-to-target to rapidly converge to near zero within approximately $5$ seconds across all tested initial AoAs, demonstrating exceptional convergence speed and consistency. The narrow shaded variance areas across multiple replicates further confirm the high reliability of the controller. \cref{fig:aoa}B presents the corresponding two-dimensional swimming trajectories within the horizontal plane. Regardless of whether the robot starts facing directly against, parallel to, or perpendicular to the background flow, the policy successfully guides the robot toward the target location (marked by the cross $\times$). The active heading adjustments (indicated by directional arrowheads) demonstrate the policy's outstanding capability for hydrodynamic adaptation, allowing the robot to utilize or counteract ambient flows effectively to maintain its position symmetrically and robustly.
}

\vspace{-0.5cm}

\subsection{Transfer to Trajectory Tracking Task}

\begin{figure*}[tb]
    \centering
    \includegraphics[width=0.99\linewidth]{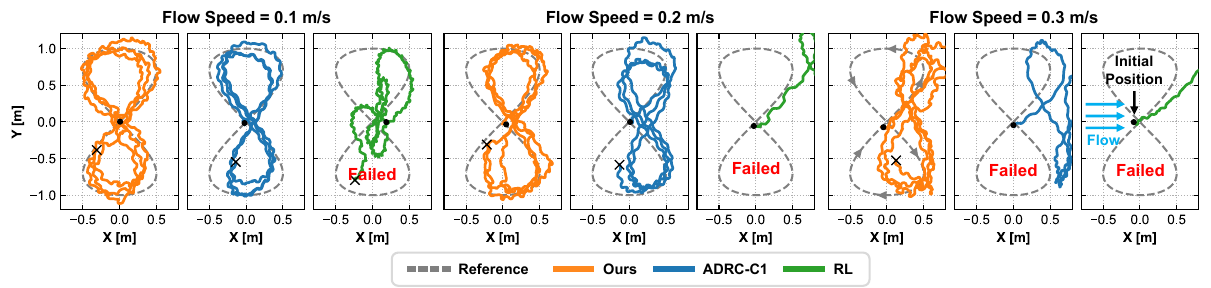}
    \vspace{-7pt}
    \caption{\color{revised}\textbf{Comparative evaluation of the generalization to trajectory tracking task in different background flows.} The robotic fish’s ability to track a challenging figure-8 trajectory is evaluated under low (0.1~m/s), medium (0.2~m/s), and high (0.3~m/s) flow velocities. The task requires the robotic fish to complete three full cycles of the figure-8 path. The results demonstrated our trained policy has superior tracking accuracy across all conditions.
    }
    \label{fig:comparison_following}
\end{figure*}

\begin{figure}[tb]
    \centering
    \includegraphics[width=0.99\linewidth]{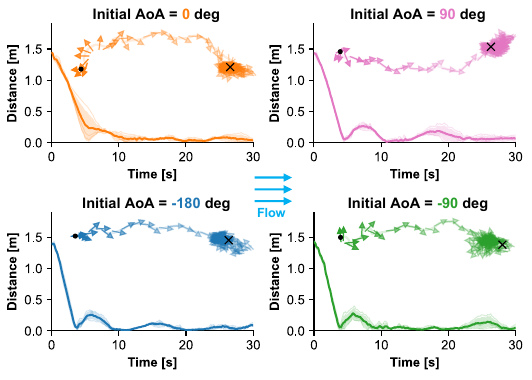}
    \vspace{-22pt}
    \caption{\color{revised}\textbf{Robustness evaluation of our trained policy under different initial angles of attack (AoA).} The time history curves of the distance-to-target and corresponding pose trajectories across 4 representative initial AoAs (i.e., -180°, -90°, 0°, and 90°) under 0.2 m/s background flow. The black dots and black crosses represent the start and end positions, respectively. The arrows represent the position and heading of the robotic fish. The results demonstrate our trained policy able to achieve fast convergence and consistent station holding regardless of the incident flow direction.}
    \label{fig:aoa}
\end{figure}

The station-holding policy developed in this study not only tackles a fundamental task in flow environments but also boasts strong transferability to other motion tasks. This is due to the fact that our policy integrates not only flow resistance mitigation for station-keeping but also target-approaching capabilities. To validate this generalization ability, we transfer the trained policy to a figure-8 trajectory tracking task and perform a comparative evaluation. In this task, the target point moves continuously along a predefined figure-8 path at a constant speed of 0.3~m/s, instead of being stationary. The trajectory is defined as
\begin{equation}
\begin{cases}
x(t) = \sin(t)\cos(t),\\[4pt]
y(t) = \sin(t),
\end{cases}
\label{eq:figure8}
\end{equation}
Experiments are conducted under three distinct background flow speeds. We compare our method with two widely adopted strategies in the robotic fish community: the PNL method \cite{guelman2007qualitative} and the ADRC-augmented method \cite{li2019bottom}. The results are summarized in \cref{fig:comparison_following}.

It can be seen in \cref{fig:comparison_following} that, at a low flow speed of 0.1~m/s, both our method and ADRC deliver strong tracking performance, while the PNL trajectory exhibits initial deformation. When the flow velocity increases to 0.2~m/s, the PNL trajectory deteriorates markedly, and the ADRC trajectory shows substantial drift. In contrast, our method still maintains a relatively complete and precise trajectory. Under a high flow speed of 0.3~m/s, both PNL and ADRC depart entirely from the target trajectory, whereas our method retains an approximate yet recognizable figure-8 shape. This is particularly striking considering the policy was trained exclusively with stationary targets. It is worth clarifying that the PNL strategy is included here to serve as a baseline for quantifying the performance improvement provided by ADRC, which treats background flow as an external disturbance. This also explains why PNL is excluded from the station-holding comparison in \cref{sec:results}.C. While ADRC’s disturbance rejection mechanism yields certain improvements, the results confirm that it still lags significantly behind our learned policy in terms of tracking accuracy and robustness under realistic flow conditions.

\vspace{-0.5cm}

\subsection{Ablation Study}

\begin{figure}[tb]
    \centering
    \includegraphics[width=0.9\linewidth]{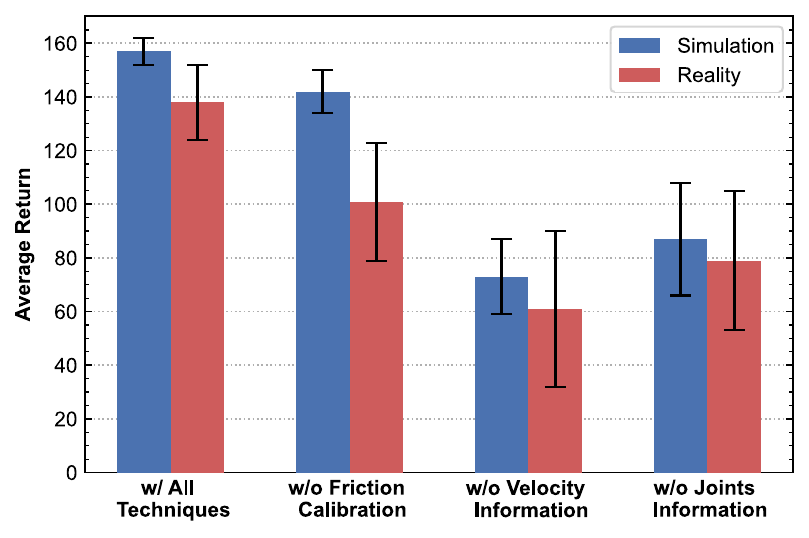}
    \vspace{-10pt}
    \caption{\textbf{Ablation evaluation of the proposed method.} For each policy, 3 trials are made in both downstream and upstream inclined at 45°, and the average and standard deviation of the cumulative rewards for each test were recorded. Each policy is trained using only data from simulation and is directly transfered to the real-world without fine-tuning.}
    \label{fig:ablation}
\end{figure}

To further validate the effectiveness of each component in our framework, we performed ablation experiments to assess three key elements---namely \emph{friction calibration}, \emph{self-velocity information}, and \emph{joints information}. All other configurations were kept consistent during training, with only these components excluded as appropriate. Consistent with the experimental setting in \cref{fig:comparison_quantitative}, station-holding tests were conducted across both simulated and real-world environments. Each test was repeated three times, and the mean and variance of cumulative rewards across 24 tests were computed, as presented in \cref{fig:ablation}. It can be observed that across both simulation and real-world settings, the method incorporating all three components achieves the best performance. The strategy excluding friction calibration still yields strong results in simulation; however, due to mismatches with real-world dynamics, its real-world performance degrades significantly. For strategies excluding self-velocity and joint information, performance decreases significantly in both simulation and real-world environments, indicating that these types of information are critical to the baseline performance of the strategy.

\begin{figure*}[t]
    \centering
    \includegraphics[width=0.99\linewidth]{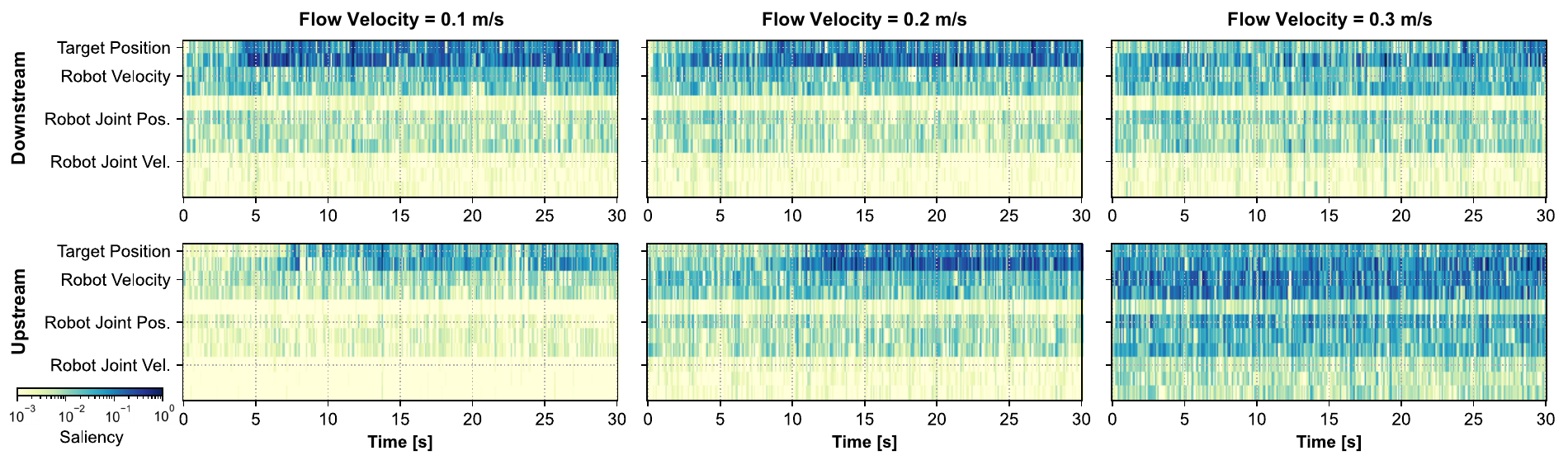}
    \vspace{-10pt}
    \caption{\textbf{Saliency analysis of state contributions under different task conditions.} Saliency maps illustrate the relative importance of input states for the learned station-holding controller across two typical task scenarios (downstream and upstream) and three flow velocities (0.1, 0.2, 0.3~m/s). Each column corresponds to a distinct flow velocity, while each row represents a task direction. Darker regions indicate higher saliency, signifying that the certain input state (e.g., target position, robot velocity, joint position, or joint velocity) has a stronger influence on the control output at corresponding time instant.}
    \label{fig:saliency}
\end{figure*}

\begin{figure}[tb]
    \centering
    \includegraphics[width=0.99\linewidth]{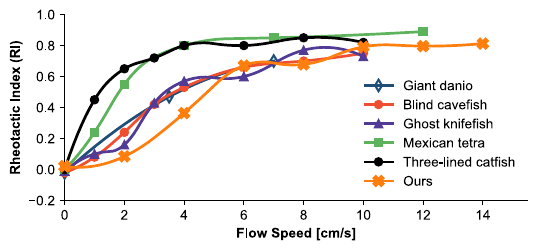}
    \vspace{-22pt}
    \caption{\color{revised}\textbf{Quantitative evaluation of the biological consistency of our trained policy.} The rheotaxis strength of our robotic fish (controlled by our trained policy) is evaluated at flow speeds from 0 to 14~cm/s and compared against biological benchmarks compiled by Coombs et al. \cite{coombs2020rheotaxis} across five fish species via rheotactic index (RI). As flow speed increases, the RI of the robotic fish rises rapidly and plateaus between 0.7 and 0.9, demonstrating an overall behavioral trend highly consistent with real fish species.}
    \label{fig:comparison_bio}
\end{figure}

\section{Discussions}\label{sec:discussion}

\subsection{Mechanistic Insights into Our Station-Holding Policy}

The unique characteristics of robotic fish sustain widespread attention across areas of  mechanical design, control, navigation and flow field perception. However, replicating \emph{rheotaxis}, a fundamental capability of biological fish (see \cref{fig:hook}), in robotic systems remains a key challenge \cite{liu2025artificial}. Traditional research \cite{zhang2015distributed,wang2024investigation} typically depends on predefined global coordinate systems, limiting the deployability of such strategies in outdoor environments, where all information must be interpreted within the robot’s local reference frame, without prior knowledge of flow velocity. 
Moreover, biological research has demonstrated that  lateral lines is not necessary for fish to maintain its position in complex ambient flows \cite{liao2006role,afridi2025beyond}. For example, James C. Liao \cite{liao2006role} experimentally showed that biological fish retain rheotactic ability even following lateral line disruption. This provides a novel perspective for position maintenance in flowing water: \emph{Can robust station holding be achieved solely using the intrinsic motion information of the robot, without explicit flow sensing devices?}

The success of our proposed policy provides a viable answer. We demonstrate that using only \textcolor{revised}{egocentric} state information, without dedicated flow sensing, the robotic fish can reliably maintain position in unknown flow fields with different velocity. Preliminary ablation analysis (see \cref{fig:ablation}) shows that this capability stems from the policy’s capacity to implicitly infer from background flow effects through variations in its own body states, effectively converting dynamic interactions into actionable feedback. To further explore this mechanism, we conducted a saliency analysis of the learned policy, as presented in \cref{fig:saliency}. The results reveal that both task direction and flow velocity will influence the importance of the state in the policy: in downstream and low-speed flows, the controller predominantly relies on target position feedback to maintain accurate holding; under stronger and upstream flows, the \textcolor{revised}{egocentric} states grow substantially in importance. This saliency analysis indicates that the learned controller leverages flow-induced perturbations as implicit sensory cues, enabling robust, anticipatory adjustments to external disturbances. This reveals a key strength of RL: the ability to capture complex hydrodynamic couplings and harness them for control, even without explicit flow measurements.

Moreover, these observations shed light on the limitations of existing controllers: their inability to compensate for or leverage background flow stems from insufficient utilization of \textcolor{revised}{egocentric} feedback. This emergent reliance on self-motion cues and hydrodynamic feedback closely mirrors biological observations of fish lateral line sensing and flow-mediated control \cite{liao2006role}. This striking convergence between biological insights and robotic implementation highlights the effectiveness of our framework in leveraging egocentric information for flow-adaptive locomotion---an approach previously underexplored and frequently deemed infeasible in robotic systems. Collectively, these results show that our learned controller not only delivers robust station-holding performance but also embodies principles consistent with natural sensorimotor coordination, bridging the divide between bio-inspired theory and embodied robotic implementation.

\textcolor{revised}{
To further provide a rigorous, quantitative validation of this biological consistency, we evaluate the station-holding behavior using the rheotactic index (RI), a metric established in biological reviews \cite{coombs2020rheotaxis,bak2013spatiotemporal} to characterize orientation strength and polarity against current, as shown in \cref{fig:comparison_bio}. The RI ranges from $-1$ (perfect negative rheotaxis) to $+1$ (perfect positive rheotaxis), with values near zero denoting no directional preference. We systematically tested the robotic fish across flow velocities ranging from $0$ to $0.14$~m/s, isolating the station-holding phase for a duration of $60$~s per trial. The experimental profiles reveal that the overall trend of the robotic fish's RI as a function of flow speed is highly congruent with compiled biological observations from real fish \cite{coombs2020rheotaxis}. Specifically, as the flow velocity increases, the robot's RI rises sharply before plateauing within a stable range of $0.7$--$0.9$, effectively mirroring the natural behavior seen across multiple aquatic species. Minor variations in finer trajectories are expected, as biological fish operate under multi-objective survival goals (e.g., foraging and predator avoidance) with internal motivational states, whereas our learned policy optimizes solely for position-holding precision and energetic efficiency. Nevertheless, this quantitative agreement strongly underscores the high biological relevance of our converged flow-adaptive control strategy.
}

\subsection{Significance and Uniqueness of the Proposed Solution}
This work has replicated a fundamental capability of biological fish in robotic platforms: \emph{station holding}. Successfully addressing this capability is critical for future deploying robotic fish in natural aquatic environments (e.g., rivers and oceans), where unknown and turbulent background flows are ubiquitous, enabling applications such as environmental monitoring and underwater inspection.
The problem exhibits exceptional complexity due to its multifaceted nature. First, the core dynamics, encompassing interactions between the robot’s oscillatory body and the surrounding fluid, are highly nonlinear even in still water. Background flow further exacerbates this nonlinearity, introducing additional forces coupled with the robot’s geometry, orientation, and flow speed. These factors form a time-varying, high-dimensional system that defies precise modeling.
Second, from a control perspective, the system is underactuated, lacking direct actuation for lateral or backward propulsion, while the flow acts as a persistent disturbance. This demands agile, multimodal swimming strategies for effective disturbance rejection. Moreover, the robot relies solely on \textcolor{revised}{egocentric} sensing, with no access to explicit flow measurements. This renders conventional model-based or feedforward control infeasible, requiring the robot to infer compensatory actions in real time through environmental interaction. These challenges make the problem notably more demanding than prior work focused on quiescent water.

Existing research often employs constrained experimental setups (e.g., narrow water channels), where controllers typically target limited degrees of freedom (e.g., longitudinal or heading control). Such simplifications overlook full positional regulation under realistic flow disturbances, severely restricting real-world applicability. This highlights the first critical gap: the absence of an experimental platform enabling free-swimming studies under tunable background flows.
Yet, an experimental platform alone is insufficient. The nonlinear, time-varying dynamics render classical model-based control methods inadequate. While RL is a promising model-free alternative, it requires extensive interaction data, which is prohibitively costly to collect in physical experiments. Most existing RL approaches rely on oversimplified models that fail to capture complex flow-induced hydrodynamics. Conventional CFD-based simulation offers high fidelity, but the high computational cost of those solvers has limited its adoption in this field. 

{\color{revised}
These challenges are specifically addressed by our proposed SWiFT framework. We identify such a unique pathway for efficiently generating complex sim-to-real control policies for robotic fish as our core methodological contribution. After comprehensive exploration of alternative approaches, our work ultimately converges on the integration of a novel high-performance GPU-based low-numerical-dissipation CFD solver with reinforcement learning (RL), calibrated against physical experiments under nominal operating conditions. This integrated framework streamlines the entire control derivation pipeline and completely bypasses the intricate workflows of conventional methods. Compared with traditional reduced-order models and many existing CFD solvers that exhibit substantially higher numerical dissipation, this low-numerical-dissipation CFD solver maintains high physical consistency even at coarse grid resolutions. When deployed as the environment for RL exploration, it enables efficient, straightforward acquisition of transferable control policies with limited dedicated measures to calibrate, and achieve effective control on physical robotic fish.

It is worth noting that while the experimental validation in this study focuses on a body and/or caudal fin (BCF) propulsion paradigm, the architectural design of the SWiFT framework is fundamentally locomotion-agnostic. By leveraging an end-to-end model-free reinforcement learning pipeline alongside fluid-structure interaction simulations, the framework can, in principle, generalize to diverse morphology and aquatic propulsion modes, including median and/or paired fin (MPF) undulation. However, extending the framework to different locomotor paradigms introduces distinct challenges. The actuator dynamics, structural compliance, and precise hydrodynamic mechanisms of MPF robots differ significantly from those of BCF systems, necessitating dedicated sim-to-real calibration efforts. Furthermore, achieving optimal policy migration across heterogeneous platforms heavily depends on the tailored design of state-action spaces and reward formulations. Consequently, while the specific configuration implemented here serves as a reference methodology, the underlying egocentric representation and training pipeline establish a transferable foundation for broader biomimetic robotic fish architectures.
}

\subsection{Limitations and Potential Application Significance}

Despite the demonstrated efficacy, several limitations of the current work point to promising avenues for future research. Firstly, the background flows generated in our physical experiments, while spatially heterogeneous, are limited to a velocity range of $0.1$--$0.3$~m/s and are primarily steady or slowly varying over a single episode. Testing in higher flow speeds or more extreme, strongly unsteady turbulent flow regimes (e.g., large vortex streets) would further validate the policy’s robustness. 

\textcolor{revised}{
It is important to recognize that as the flow speed approaches the maximum sustained swimming capability of the robotic fish, the primary challenge transitions from algorithmic control to physical constraints, where the available thrust must be predominantly diverted to counteract the incoming flow, leaving a diminishing control authority for active disturbance rejection. To experimentally characterize the complexity of the current flow field beyond idealized laminar flows, we conducted 21 repeated trials tracking a lightweight floating marker, as shown in \cref{fig:flow}. The marker underwent significant lateral displacement and rapid rotations during its upstream-to-downstream transit, quantitatively substantiating that the evaluation environment possesses non-negligible turbulent characteristics. When encountering even more severe, unmodeled outdoor turbulence, our egocentric learning paradigm is inherently well-suited to generalize. Because the policy relies purely on onboard body-frame sensory feedback rather than macroscopic global flow vectors, the same learning pipeline can be scaled to extreme environments simply by augmenting the simulation training distribution with high-fidelity turbulence models, provided the perturbations do not exceed the absolute physical bounds of the robot's propulsive capacity.
}

Secondly, our method relies on robust perception of the target position and accurate estimation of self-states (linear and angular velocity). In practical outdoor deployment, however, recognizing and localizing target references remain challenging due to adverse visual conditions, such as turbidity, variable illumination, or occlusions, in natural aquatic settings. Thirdly, the proposed framework exhibits limited performance under low-speed flow regimes. Unlike in high-speed flow conditions, where deviations from the target position elicit timely feedback for corrective actions, disturbances in low-flow environments are more subtle and less perceptible from \textcolor{revised}{egocentric} states, often leading to delayed feedback control.

\textcolor{revised}{
Fourthly, while the present evaluation focuses on planar (2D) maneuvers within a flow tank, open-water deployments inevitably subject the robot to complex three-dimensional flow structures. Extending the SWiFT framework to full 3D attitude and motion control introduces critical challenges across multiple dimensions. From an algorithmic perspective, our CFD-based simulator natively supports 3D fluid-structure interactions and rigid-body dynamics, and the model-free RL pipeline is inherently agnostic to state-action dimensionalities, making the 3D scaling conceptually straightforward. However, achieving fully autonomous 3D deployment demands substantial hardware and sensing upgrades. Onboard sensory modalities must be expanded with depth sensors and inertial measurement units (IMUs) to reconstruct the full 3D egocentric state under localized vortex disturbances. Concurrently, the mechanical platform requires supplementary actuators, such as pectoral fins or buoyancy regulation mechanisms, to govern additional degrees of freedom (depth, roll, and pitch), which expands the action exploration space during reinforcement learning. Addressing these coupled sensing and actuation scaling challenges represents an essential direction for our future work.
}

With all these considerations, this work holds considerable practical significance. By achieving robust station holding in unknown dynamic flows, it directly enables long-duration missions for robotic fish in real-world environments. This capability is foundational for applications including ecological monitoring at fixed river locations, inspection of current-exposed underwater infrastructure (e.g., bridge piers, pipeline junctions), and position maintenance within fish schools for biological research. Finally, the proposed framework, along with its demonstrated resilience to simulation abstractions during control transfer, provides a powerful tool for the computational exploration of effective controllers for other fluid-interactive systems (e.g., ornithopters) in complex hydrodynamic settings.

\section{Conclusions}\label{sec:conclusion}

In this work, we address one of the most fundamental yet unresolved challenges in robotic fish control, maintaining position in unknown and dynamic background flows, referred to as ``station holding''.
To tackle this challenge, we developed SWiFT, an integrated framework encompassing a low numerical dissipation CFD-based simulator, a systematic sim-to-real calibration pipeline, and a flow tank for free-swimming experiments. 

\textcolor{revised}{
By co-designing a low numerical dissipation LBM with an end-to-end RL architecture, SWiFT overcomes the historical computational barrier between  fluid simulation and data-intensive learning. This integration enables, for the first time, the systematic formulation and experimental validation of egocentric station holding for a freely swimming robotic fish. Crucially, we demonstrate that egocentric feedback alone provides sufficient information for robust flow rejection, bridging biological observations with engineering practice for low-cost autonomous underwater robots.}




\bibliographystyle{IEEEtran}
\bibliography{refs}



\end{document}